\PassOptionsToPackage{unicode}{hyperref}
\PassOptionsToPackage{hyphens}{url}
\PassOptionsToPackage{dvipsnames,svgnames,x11names}{xcolor}
\documentclass[
  11pt,
]{article}
\usepackage{xcolor}
\usepackage[margin=1in]{geometry}
\usepackage{amsmath,amssymb}
\usepackage{iftex}
\ifPDFTeX
  \usepackage[T1]{fontenc}
  \usepackage[utf8]{inputenc}
  \usepackage{textcomp} 
\else 
  \usepackage{unicode-math} 
  \defaultfontfeatures{Scale=MatchLowercase}
  \defaultfontfeatures[\rmfamily]{Ligatures=TeX,Scale=1}
\fi
\usepackage{lmodern}
\ifPDFTeX\else
\fi
\IfFileExists{upquote.sty}{\usepackage{upquote}}{}
\IfFileExists{microtype.sty}{
  \usepackage[]{microtype}
  \UseMicrotypeSet[protrusion]{basicmath} 
}{}
\makeatletter
\@ifundefined{KOMAClassName}{
  \IfFileExists{parskip.sty}{%
    \usepackage{parskip}
  }{
    \setlength{\parindent}{0pt}
    \setlength{\parskip}{6pt plus 2pt minus 1pt}}
}{
  \KOMAoptions{parskip=half}}
\makeatother
\usepackage{longtable,booktabs,array}
\usepackage{calc} 
\usepackage{etoolbox}
\makeatletter
\patchcmd\longtable{\par}{\if@noskipsec\mbox{}\fi\par}{}{}
\makeatother
\IfFileExists{footnotehyper.sty}{\usepackage{footnotehyper}}{\usepackage{footnote}}
\makesavenoteenv{longtable}
\usepackage{graphicx}
\makeatletter
\newsavebox\pandoc@box
\newcommand*\pandocbounded[1]{
  \sbox\pandoc@box{#1}%
  \Gscale@div\@tempa{\textheight}{\dimexpr\ht\pandoc@box+\dp\pandoc@box\relax}%
  \Gscale@div\@tempb{\linewidth}{\wd\pandoc@box}%
  \ifdim\@tempb\p@<\@tempa\p@\let\@tempa\@tempb\fi
  \ifdim\@tempa\p@<\p@\scalebox{\@tempa}{\usebox\pandoc@box}%
  \else\usebox{\pandoc@box}%
  \fi%
}
\def\fps@figure{htbp}
\makeatother
\usepackage{bookmark}
\IfFileExists{xurl.sty}{\usepackage{xurl}}{} 
\hypersetup{
  colorlinks=true,
  linkcolor={Maroon},
  filecolor={Maroon},
  citecolor={Blue},
  urlcolor={Blue},
  pdfcreator={LaTeX via pandoc}}

\author{}
\date{}

\begin{document}

\section{When Algorithms Meet Artists: Semantic Compression and
Stakeholder Marginalisation in Public AI-Art Discourse
(2013-2025)}\label{when-algorithms-meet-artists-semantic-compression-and-stakeholder-marginalisation-in-public-ai-art-discourse-2013-2025}

\textbf{Authors:} Ariya Mukherjee-Gandhi* and Oliver
Muellerklein\textsuperscript{1,*,†}

\textsuperscript{1} University of California, Berkeley\\
* These authors contributed equally to this work\\
† Corresponding author: omuellerklein@berkeley.edu

\emph{Preprint, July 2026 (arXiv:2508.03037v5). This version supersedes
the February 2026 preprint: the analysis pipeline was rebuilt and all
quantitative results updated.}

\subsection{Abstract}\label{abstract}

Artists occupy a paradoxical position in generative AI. Their own work
trains models that now compete with them, replicate their styles, and
reshape the creative economy they inhabit. Yet whether artist concerns
achieve proportional representation in the public discourse that shapes
AI governance remains an open empirical question. We mapped the semantic
landscape of public AI-art discourse from 2013 to 2025, drawing on 1,736
text chunks from news, podcasts, legal filings, and research, and
projected 252 US-based practising artists' survey responses, captured
across 70 unique frames spanning five concern dimensions, into the same
space. We identify what we term \emph{semantic compression}, the
systematic narrowing of a diverse set of stakeholder concerns into a
narrow region of public meaning-space. Compression is selective. Nearly
all artist statements concentrate in just two of twenty discourse
topics, while most of the remaining discourse volume sits in topics with
no artist voice at all. Public discourse speaks about artists in regions
where artists themselves are absent. Compression also operates inside
the topics that do contain artists: dozens of distinct ownership and
utility positions flow into a single topic framed as a debate about
aesthetic authenticity, persisting after we control for differences in
writing style between surveys and media. Public discourse amplifies
broad tropes about creativity while attenuating the specific, actionable
regulatory claims that artists make. Amid ongoing coalition mobilisation
and litigation over AI training, these findings identify a structural
pattern by which primary stakeholders are rendered peripheral in the
very discourse that shapes their conditions of practice.

\textbf{Keywords:} stakeholder representation, semantic compression,
epistemic marginalisation, AI governance, topic modelling, generative
AI, AI social science, sociotechnical systems

\subsection{Introduction}\label{introduction}

\subsubsection{Stakeholder positions and public
discourse}\label{stakeholder-positions-and-public-discourse}

Public discourse about emerging technologies shapes which stakeholder
concerns achieve regulatory salience and which are rendered peripheral.
As generative AI reshapes the production, circulation, and valuation of
creative work, the question of whose concerns become legible to
policymakers is one of both communicative justice and policy design. We
investigate this question in a consequential domain: generative AI and
creative labour.

Artists occupy a paradoxical position in this technological transition.
They stand as both the primary authors of training data for these
systems and the stakeholders most susceptible to the systems' disruptive
impacts. Generative AI models are intensifying persistent disputes
concerning consent, compensation, authorship, and the legitimacy of data
practices (Lovato et al., 2024; Creative Rights in AI Coalition, 2024;
Creators Coalition on AI, 2025). The effects of these disputes have
become increasingly tangible. In 2023, a landmark class-action lawsuit
(\emph{Andersen v. Stability AI}) brought artists' grievances before a
federal court, alleging that generative image models developed using
copyrighted works without consent constitute mass infringement (Andersen
et al.~v. Stability AI, 2023). Scholarship has begun to articulate the
normative basis for such claims, arguing that the unauthorised use of
artists' work to train commercial generative systems is best understood
as a form of labour theft rather than a technical edge case of copyright
doctrine (Goetze, 2024). As Varvasovszky and Brugha (2000) emphasise,
identifying and understanding primary stakeholders is essential for
evaluating the social consequences of technological change. For artists,
this transformation raises fundamental questions about creative
ownership, labour, and artistic identity (McCormack et al., 2019;
Browne, 2022).

Relevant research shows that we cannot treat ``artists'' as a single,
unified group in their interpretation of AI developments. Studies by
Lovato et al.~(2024) and Kawakami and Venkatagiri (2024) reveal that
artist outlooks are actually highly specific and often internally
complex. Lovato et al.~(2024) operationalise these concerns across five
dimensions (Utility, Ownership, Compensation, Transparency, and Threat)
comprising 70 unique viewpoints (\emph{i.e.~frames}), understood here in
Entman's (1993) sense as interpretive packages that select and emphasise
particular aspects of a perceived reality. This granularity is
essential. One artist may embrace AI tools while rejecting the economic
models of companies that develop them. Another might support
transparency requirements but remain uncertain about ownership claims.
Kawakami and Venkatagiri (2024) document this spectrum, from workflow
augmentation benefits to considerable employment risks, and these
concerns differ across artistic communities (Jiang et al., 2023). The
complexity extends to public perception. When artwork is labelled as
AI-generated, aesthetic appreciation drops regardless of quality,
suggesting that the ``AI art'' label itself functions as a compressive
frame that flattens the diversity of human-AI creative interactions
(Małecki et al., 2025).

These dimensions matter not only because they document disagreement
among artists, but also because they provide the empirical basis to
define \emph{stakeholder frames}, coherent interpretive positions that
can be compared with the accounts that dominate broader public
discourse.

The internal structure of these frames defies the binary categories that
dominate public coverage. Table 1 presents four representative profiles
from our sample. The first, which we call the ``Pragmatic Dual-Holder'',
views AI as both a threat to employment and a useful creative tool,
favours mandatory disclosure, asserts artist ownership, yet would donate
their work freely as training data. This position is internally coherent
(one can recognise a danger while also seeing practical value) but the
``Threat versus Tool'' framing of media coverage cannot accommodate it.
Of the 252 artists in our sample, only 104 (41.3\%) fit cleanly into
either pole of that binary: 60 who view AI as purely threatening and 44
who view it as purely positive. The remaining 148 (58.7\%) hold
positions the binary cannot represent, including 55 (21.8\%) who agree
that AI is both a threat and a positive development, a finding
consistent with Lovato et al.'s (2024) observation that 22\% of their
full sample held what they termed ``complex opinions'' spanning both
threat and utility.

Free-text responses from the same survey illustrate the specificity of
these positions. One artist stated, ``The artist. Period. This
technology is trained using copyrighted material. They have no right to
profit from it.'' Another offered a more uncertain view: ``I'm not sure
yet. I do think that if you are using someone else's work, there should
be a financial contribution. I'm still learning about this.'' A third
proposed a concrete policy mechanism: ``Companies and individuals should
be subject to a new tax in order to use AI art that was trained with
your artwork.'' These are distinct governance demands that require
different policy responses.

The diversity extends well beyond the threat-utility axis. Across all
five dimensions, artists produced 137 unique cross-dimensional frame
combinations from 252 respondents. The most common single combination
accounts for just 7.1\% of the sample. Artists who have used AI art
models firsthand are nearly three times more likely to hold the dual
threat-and-utility position (29.7\%) than those who have not (10.9\%;
\(\chi\)\textsuperscript{2} = 11.31, p \textless{} 0.001, Cramer's V =
0.21), suggesting that direct experience with the technology produces
more nuanced positions, not simpler ones.

\textbf{Table 1: Representative artist frame profiles (n = 252 US-based
practising artists)}

\begin{longtable}[]{@{}
  >{\raggedright\arraybackslash}p{(\linewidth - 12\tabcolsep) * \real{0.2034}}
  >{\raggedright\arraybackslash}p{(\linewidth - 12\tabcolsep) * \real{0.1102}}
  >{\raggedright\arraybackslash}p{(\linewidth - 12\tabcolsep) * \real{0.1186}}
  >{\raggedright\arraybackslash}p{(\linewidth - 12\tabcolsep) * \real{0.1610}}
  >{\raggedright\arraybackslash}p{(\linewidth - 12\tabcolsep) * \real{0.1356}}
  >{\raggedright\arraybackslash}p{(\linewidth - 12\tabcolsep) * \real{0.1864}}
  >{\raggedright\arraybackslash}p{(\linewidth - 12\tabcolsep) * \real{0.0678}}@{}}
\toprule\noalign{}
\begin{minipage}[b]{\linewidth}\raggedright
\textbf{Profile}
\end{minipage} & \begin{minipage}[b]{\linewidth}\raggedright
\textbf{Threat}
\end{minipage} & \begin{minipage}[b]{\linewidth}\raggedright
\textbf{Utility}
\end{minipage} & \begin{minipage}[b]{\linewidth}\raggedright
\textbf{Transparency}
\end{minipage} & \begin{minipage}[b]{\linewidth}\raggedright
\textbf{Ownership}
\end{minipage} & \begin{minipage}[b]{\linewidth}\raggedright
\textbf{Compensation}
\end{minipage} & \begin{minipage}[b]{\linewidth}\raggedright
\textbf{n}
\end{minipage} \\
\midrule\noalign{}
\endhead
\bottomrule\noalign{}
\endlastfoot
Pragmatic Dual-Holder & Agree & Agree & Pro-disclosure & Artist owns &
Would donate freely & 18 \\
Protective Advocate & Agree & Disagree & Pro-disclosure & Artist owns &
Revenue share & 7 \\
Open-Source Embracer & Disagree & Agree & Anti-disclosure & Does not own
& Would donate freely & 10 \\
Cautious Observer & Neutral & Neutral & Pro-disclosure & Does not own &
Revenue share & 14 \\
\end{longtable}

\emph{Profiles (author generated labels) represent the most common
combination within each threat-utility position. Full dimension
distributions are in Supplementary Table S4.}

However, public discourse often fails to capture this heterogeneity and
representation is not neutral. Crawford (2021) argues that AI
classification systems perform politics by deciding whose categories
count, rendering certain knowledge legible and other knowledge residual
or invisible. As media and communication scholarship has established,
public discourse does not merely reflect reality but actively
constitutes it through selective framing, agenda-setting, and the
differential amplification of particular voices and concerns (Entman,
1993; Jasanoff, 2015). When technology discourse foregrounds certain
actors and narratives while leaving others at the periphery, the
resulting public understanding becomes structurally skewed.

Public AI discourse gives disproportionate standing to economic and
scientific actors, foregrounding narratives of ``progress'' and economic
potential while overshadowing ethical and social frames (Zai et al.,
2025). In AI-art coverage specifically, news media tend to reduce the
complex interface between AI and creative practice into binary
oppositions (``AI as Threat'' versus ``AI as Tool''), stripping away the
nuanced labour concerns held by practitioners (Bøgh, 2025). This pattern
extends to public debates about image-generative AI and Large Language
Models, where ten distinct thematic topics exist but particular framings
achieve dominance while others remain peripheral (Banks and Li, 2025).
It also appears in media coverage of automation more broadly, where
``responsibility networks'' assign agency to institutional actors
(\emph{e.g.~curators, platform operators, AI companies, regulators})
while marginalising those most affected (Saurwein et al., 2025).

The structural mechanics of the contemporary public sphere exacerbate
this phenomenon. Traditional editorial gatekeeping has given way to
``algorithmic agenda-setting,'' where recommender systems prioritise
content that maximises engagement rather than epistemic completeness
(Sichach, 2024). In this ``post-mediatized'' sphere, algorithmic
curation establishes the epistemic boundaries of discourse, determining
which voices are amplified and which remain invisible (Pane, 2025). The
pattern recurs across platforms and geographies. In Chinese AI-education
coverage, government and corporate voices dominate while students and
teachers receive minimal representation (Huang and Gadavanij, 2025), and
in an analysis of 368,000 ChatGPT-related tweets, ethical concerns
circulate but remain peripheral to narratives of capability and novelty
(Cohen et al., 2026). The evidence suggests that the social and
structural conditions exist for the marginalisation of artist voices.
This hypothesis has not been empirically tested in the AI-art domain
using computational methods that can quantify the degree and nature of
this marginalisation.

\subsubsection{Epistemic marginalisation as a measurable
outcome}\label{epistemic-marginalisation-as-a-measurable-outcome}

Recent scholarship on epistemic injustice in AI governance provides a
useful theoretical lens. Building on Fricker's (2007) framework, in
which testimonial injustice arises when prejudice diminishes a speaker's
credibility and hermeneutical injustice arises when the collective
interpretive resources available to a community cannot accommodate the
experience of its marginalised members, Kay, Kasirzadeh and Mohamed
(2024) extend the framework to generative AI specifically, identifying
four dimensions of \emph{generative algorithmic epistemic injustice}:
amplified testimonial injustice, manipulative testimonial injustice,
hermeneutical ignorance, and access injustice. This framing is
consistent with relational accounts of algorithmic injustice that locate
harm not in individual model outputs but in the structured asymmetries
between those who build systems and those subject to them, and with the
corresponding argument that responses focused exclusively on technical
correctives fail to centre the disproportionately impacted communities
themselves (Birhane, 2021).

When knowledge produced by affected groups is backgrounded relative to
institutional or technical narratives, marginalisation can occur without
explicit exclusion. It operates through differential amplification,
topic selection, and standing asymmetries rather than overt silencing, a
form of structural injustice that the present analysis is positioned to
measure directly in public discourse. In the AI-art debate, where
artists are directly affected by data practices and market
restructuring, epistemic marginalisation is a plausible outcome of the
visibility regimes documented by Sichach (2024) and Pane (2025). The
question is not whether artists are formally prohibited from speaking,
but whether their concerns achieve adequate salience within the
discursive ecosystem that shapes public understanding and policy
attention. This framing connects to broader questions in AI social
science about how AI enters and reshapes social worlds, and how those
worlds, in turn, shape AI (Joyce and Cruz, 2024).

The timing of this question matters because AI-art discourse has shifted
sharply over the last decade. Early public discussion (2013 to 2020)
disproportionately emphasised novelty, experimentation, and
philosophical speculation about machine creativity. Milestones such as
Google's DeepDream (Mordvintsev et al., 2015) and the emergence of
Generative Adversarial Networks (Goodfellow et al., 2014) spawned
discussions of algorithmic aesthetics and machine consciousness. This
period generated substantial scholarly attention to the ontological
status of AI-generated works and their place in art-historical lineages
(Browne, 2022; Mazzone and Elgammal, 2019). The 2018 Christie's auction
of \emph{Portrait of Edmond de Belamy}, a GAN-generated image, brought
AI art into mainstream cultural discourse, intensifying debates about
authorship and market legitimacy.

The more recent period (2021 to 2025), characterised by the mass
adoption of text-to-image systems like DALL-E (Ramesh et al., 2021) and
latent diffusion models (Rombach et al., 2022), has seen a pivot toward
labour disruption, legal conflict, and institutional integration. This
pivot has been accompanied by a shift in media framing. Coverage
oscillates between ``opportunity'' and ``crisis'' narratives,
simplifying stakeholder realities into repeatable tropes (Cheung, 2025).
Topic modelling studies confirm that particular themes achieve
prominence at specific temporal moments, as in generative art discourse
(Saeedi and Taleghani, 2025) and in longitudinal coverage of AI virtual
assistants, where discourse moved from personal impact to societal
concern frames as the technology matured (Wald et al., 2026). However,
these analyses focus on the internal structure of public discourse
without comparing it to empirically-derived stakeholder positions.
Despite this shift, we lack empirical evidence on the alignment between
public narratives and stakeholder priorities. Does public discourse
reflect the distribution of artists' concerns, or does it compress them
into a narrow semantic region?

This gap is especially important for media and communication research
because it connects a normative claim (that artists are primary
stakeholders whose voices should inform AI governance) to a measurable
outcome: representation and salience within public discourse. It
connects to broader questions about how public attention is allocated
across competing frames and how stakeholder knowledge is amplified or
attenuated in mediated discourse (Cohen et al., 2026; Wald et al.,
2026). Existing research has mapped public debates about generative AI
(Saeedi and Taleghani, 2025; Banks and Li, 2025) and documented
stakeholder perspectives separately (Kawakami and Venkatagiri, 2024;
Jiang et al., 2023; Mazzone and Elgammal, 2019), but seldom connects the
two sides empirically. Stakeholder frames have not been
\emph{geometrically projected} into public discourse space to measure
representational distribution within the broader discursive ecosystem.
As DiMaggio et al.~(2013) argue, topic modelling offers a powerful tool
for exploiting affinities between computational text analysis and the
sociological study of culture, enabling researchers to map meaning
structures at scale while remaining attentive to the relational
positions of different actors within those structures.

We still lack evidence on whether public attention tracks the
distribution of stakeholder frames or instead produces what we term
semantic compression: the collapse of a diverse set of artist concerns
into a narrow region of the public meaning space. If public discourse
systematically compresses stakeholder complexity, amplifying broad
tropes and attenuating nuanced positions, the consequences for AI
governance are significant. Policymakers, platforms, and institutions
that rely on public discourse as a proxy for stakeholder priorities
would underweight the concerns of affected communities.

\subsubsection{This study: mapping discourse, projecting
stakeholders}\label{this-study-mapping-discourse-projecting-stakeholders}

We address this gap by mapping the semantic landscape of public
discourse on AI and art (2013 to 2025) and projecting artist survey
statements into that space to quantify alignment, separation, and
semantic compression. Our public corpus spans 125 documents and 1,736
processed text chunks across news, podcasts, panel discussions, legal
filings, and research papers, yielding 20 distinct discourse topics. We
operationalise stakeholder viewpoints and generate 1,259 artist probes
capturing 70 unique frames across the five concern dimensions from 252
US-based practising artists from Lovato et al.~(2024). Three questions
structure the analysis that follows. First, whether artist frames
concentrate in a narrow region of the discourse topic space. Second,
whether any such concentration reflects substantive semantic divergence
or merely a stylistic mismatch between survey-format and media-format
text. Third, if compression is present, whether it operates through a
single mechanism or through several structurally distinct operations
acting in concert.

\subsection{Methods}\label{methods}

\subsubsection{Study design overview}\label{study-design-overview}

We measured whether artist stakeholder concerns achieve proportional
representation in public discourse about generative AI and art. Our
approach constructs a semantic reference map from public discourse (2013
to 2025) and projects artist survey responses into that shared space to
quantify representational alignment. This method builds on recent
computational work for comparing how different stakeholder groups
articulate contested topics (Elmholdt et al., 2025), and with broader
precedents for applying word-embedding methods to social-scientific
discourse analysis (Matsui and Ferrara, 2024).

\subsubsection{Data sources and corpus
construction}\label{data-sources-and-corpus-construction}

\paragraph{Public discourse corpus}\label{public-discourse-corpus}

We assembled a multimodal corpus covering 2013 to 2025 using targeted
Google Search queries related to generative AI and art (e.g., ``AI art
impact on artists,'' ``generative AI copyright'', with the full query
list available in our online repository). For each query, we sampled all
unique documents appearing on the first page of results. Searches were
conducted in a logged-out browser environment to reduce personalisation
artifacts. We acknowledge that ranking on a commercial search engine
reflects advertising economics and proprietary relevance signals rather
than any neutral measure of public salience (Noble, 2018). Our corpus
inherits that visibility hierarchy, and we treat it as a feature of the
public-discourse object we are characterising rather than as an unbiased
sample of all possible discourse.

We included English-language sources from news articles, blogs,
podcasts, interviews, panel discussions, peer-reviewed papers, and legal
filings. We excluded social-media-first platforms (Reddit, Twitter/X) to
maintain comparability with search-ranked and editorialised sources.
Recent research raises further concerns about the authenticity of social
media discourse as a proxy for public opinion. Li et al.~(2024) found
that approximately 47\% of accounts in activist discourse on Twitter/X
were bots, and that bot interaction systematically distorted the
sentiment of human participants. Spoken sources were transcribed using
OpenAI Whisper with manual review.

The final corpus comprised 125 documents segmented into 1,736 text
chunks (\textasciitilde250 words each with 25-word overlap). Each chunk
was annotated with publication year and media type for stratified
robustness testing.

\paragraph{Artist survey corpus}\label{artist-survey-corpus}

Artist stakeholder language was derived from the survey dataset of
Lovato et al.~(2024), which captured practising artists' attitudes
toward generative AI. We filtered to US-based respondents actively
practising as artists (n = 252) to match the geographic scope of our
public corpus. The sample spans multiple creative practices, including
painting (n=108), photography (n=49), writing (n=27), design (n=19),
digital practice (n=11), music (n=9), craft (n=7), illustration (n=6),
sculpture (n=4), drawing (n=3), tattoo (n=2), and other maker practice
(n=6). Visual practices predominate but are not exclusive. For example,
writers and musicians together account for \textasciitilde14\% of
respondents.

To enable direct comparison in shared semantic space, we converted
survey responses into short declarative probes using fixed templates
(e.g., ``I agree that AI art models are a threat to art workers''). This
produced 1,259 artist probes spanning the five concern dimensions. One
of the 252 respondents did not answer the threat item, yielding 251
threat probes rather than 252. All other dimensions have complete
responses. For single-question dimensions (threat, utility,
transparency), each respondent generates one of three probes (agree,
neutral, disagree). For multi-question dimensions, responses are
combined. Ownership encompasses three sub-questions on who should own
AI-generated work, producing 24 unique frames from the combinatorial
responses. Compensation includes multiple compensation models with
optional free-text elaboration, producing 37 unique frames. In total,
the 29 within-dimension response categories from Lovato et al.~expand
into 70 unique frames once sub-questions are combined. A single
practicing artist may express up to one unique frame per concern
dimension (\emph{i.e.~up to five frames per person}).

\paragraph{Public probes (style
control)}\label{public-probes-style-control}

A key confound in comparing survey statements to public discourse is
stylistic mismatch. To check whether the divergence is stylistic or
substantive, we extracted style-matched public probes, short declarative
sentences from public documents that match the syntactic form of artist
probes, retrieved via embedding similarity in the shared semantic space.

We generated 250 synthetic Likert-style anchor statements following a
factorial design (5 themes \(\times\) 5 agreement levels, strongly
disagree through strongly agree, \(\times\) 10 discourse-style variants
such as blog opinion, news editorial, artist interview, and legal
brief), then retrieved the nearest-neighbour sentences from the public
corpus for each anchor by cosine similarity over e5-large-v2 embeddings.
The 250 anchors served only as retrieval queries and were discarded
after extraction. After programmatic filtering to retain only on-theme
sentences and deduplication per theme, this yielded 750 style-matched
public probes. Retrieval in the same semantic space used for the main
analysis ensures that the style controls are genuinely style-matched
rather than lexically pre-selected, which matters because an AI-based
methodology risks circularity if the style-matched set is constructed
with keyword criteria that anticipate the themes being tested. Human
verification of the style-matched public probes was performed to ensure
validity. The full design matrix and representative examples are
available in our online repository.

\subsubsection{Semantic mapping}\label{semantic-mapping}

All text units were encoded using the sentence-transformers/e5-large-v2
embedding model (1,024 dimensions), with the ``query:'' input prefix
recommended by the model card for clustering and classification tasks.
We selected this model over smaller alternatives (e.g., all-MiniLM-L6-v2
at 384 dimensions) because our corpus spans multiple genres, registers,
and a thirteen-year temporal window. The E5 model family was trained on
heterogeneous text pairs for cross-domain similarity, making it better
suited than general-purpose sentence encoders for comparisons where the
same concept may be expressed in journalistic, legal, survey, or
academic registers. The higher dimensionality preserves finer-grained
semantic distinctions, which is critical for detecting whether
stakeholder groups occupy genuinely distinct regions of the embedding
space.

To ensure stable cross-corpus comparison, we implemented a
consensus-based approach. We generated 30 UMAP projections with distinct
random seeds (n\_neighbors = 53, min\_dist = 0.01, n\_components = 5,
cosine metric), computed the pairwise Euclidean distance matrix in 5D
for each run, and averaged these matrices to produce a consensus
distance structure. A final UMAP was fit from the consensus distance
matrix as a precomputed metric. This procedure increases average
seed-to-consensus Adjusted Rand Index from 0.56 (naive coordinate
averaging) to 0.71 (distance-matrix consensus), yielding a reference
geometry stable enough for new data to be projected without distorting
the base map. Full implementation details and stability validation are
provided in Supplementary Information and the online code repository.

We then trained a projection head to map from the 1,024-dimensional
embedding space to the 5-dimensional consensus coordinates, enabling
placement of artist probes and public probes into the reference
geometry. The projection head is a scikit-learn MLPRegressor with four
hidden layers of dimensions 1024 \(\rightarrow\) 512 \(\rightarrow\) 256
\(\rightarrow\) 128, ReLU activation, and L2 weight regularisation
(\(\alpha\) = 0.0001). Training used the Adam optimiser with an initial
learning rate of 0.002 and up to 1,000 iterations, with early stopping
triggered on a 10\% inner validation split drawn from the training set.
Inputs and outputs were standardised via StandardScaler fitted on the
training split, and the model was fit on an 85/15 train/test partition
of the 1,736 public chunks (random state 42). On the held-out 15\% test
set the projection achieves R\textsuperscript{2} = 0.904 against the
consensus coordinates. Artist probes and style-matched public probes are
never seen during fitting and are projected only after training through
the fixed MLP. Sensitivity analysis (Figure S4) confirms that the top-4
topic concentration of artist probes exceeds 91\% across all tested
cluster counts (k=10 to 30), indicating that the concentration finding
does not depend on the k=20 choice.

The 1,736 public chunks were clustered using KMeans at k=20, selected
via a four-stage hyperparameter validation across 497 configurations.
Selection balanced three criteria: consensus silhouette of 0.657 at k=20
(computed on the 5-dimensional consensus UMAP coordinates with Euclidean
distance; see Figure S3 for the full k=5-35 curve), fewest
single-article topics (3 of 20), and the highest proportion of valid
topics (85\%, where validity required at least 10 chunks drawn from at
least 2 distinct source articles). The resulting 20 topics form the
semantic inventory for all comparisons reported below. This pipeline
follows the embedding-then-cluster tradition popularised in BERTopic
(Grootendorst, 2022), with two principled departures. We substitute
consensus KMeans for HDBSCAN to obtain stable cross-corpus comparisons,
and we interpret topics via multi-annotator LLM synthesis rather than
class-based TF-IDF alone.

\subsubsection{Topic interpretation}\label{topic-interpretation}

Each topic was interpreted independently by six annotators: two human
coders and four LLMs (Claude Opus 4.6, Claude Sonnet 4.6, GPT-5.4-mini,
GPT-5.4-nano). This multi-annotator approach addresses subjectivity in
topic labelling (Ziems et al., 2024). The research team synthesised the
six independent annotations into candidate labels, keywords, and
macro-thematic groupings, and finalised all labels by consensus (see
online repository for full topic inventory).

We treat the resulting labels as panel-derived interpretations of
cluster contents rather than ground-truth descriptions: when later
sections refer to a topic "about" or "labelled as" a particular theme,
that reflects what trained readers (four LLMs and two human coders, all
working from the chunks in the cluster) converged on as the
cluster\textquotesingle s most defensible reading. The compression
findings reported below depend on the gap between this panel reading and
where artist probes actually land. That gap, not the labels per se, is
the measurement.

\subsubsection{Inference framework}\label{inference-framework}

We employed three complementary analyses to establish robustness:

\paragraph{Distributional analysis}\label{distributional-analysis}

We compared how corpora distribute across the 20 topics using chi-square
tests, Cramér's V (effect size), and Jensen-Shannon divergence. Pairwise
comparisons tested: Public vs.~Artist Probes (raw divergence), Public
Probes vs.~Artist Probes (style-controlled divergence).

\paragraph{Geometric analysis}\label{geometric-analysis}

We computed centroid distances between corpora in 5D space and
k-nearest-neighbour same-source rates. Permutation tests (10,000
iterations) generated null distributions for significance testing.

\paragraph{Robustness checks}\label{robustness-checks}

We tested whether divergence patterns could be explained by media type
(news article, blog post, podcast transcript, interview, panel
discussion, peer-reviewed paper, or legal filing) or by publication-year
bin (2013--2017, 2018--2021, 2022--2025). Within each stratum we
recomputed Cramér's V and the chi-square statistic between the artist
and public-probe distributions across the 20 topics, which verifies that
the observed divergence is not an artefact of a single source type or a
single temporal window.

Statistical tests were conducted with \(\alpha\)= 0.05 throughout. The
\(\chi^{2}\)tests of independence between corpora across the 20 topics
are two-sided. Tests of distributional divergence (Jensen-Shannon
divergence), geometric separation (centroid distance), and per-theme
entropy are non-negative magnitude statistics and use one-tailed
permutation tests with 10,000 iterations, with p-values reported as
(count + 1) / (number of permutations + 1) to avoid p = 0. Concentration
tests use the same convention against a uniform-over-20-topics null.
Inter-rater agreement between the two validation LLMs is reported as
Cohen's \(\kappa\) with 95\% Wilson confidence interval. Per-theme
recall and binomial-proportion estimates are reported with 95\% Wilson
intervals. Effect-size companions (Cramér's V for the \(\chi^{2}\)tests)
are reported alongside the test statistics.

\subsubsection{Differential compression
metrics}\label{differential-compression-metrics}

To quantify how compression varies across artist concern dimensions, we
computed three complementary metrics for each theme.

First, we measured distributional entropy across the 20 topics for each
theme. Shannon entropy captures the spread of a distribution. A theme
concentrated in a single topic has entropy near zero, while a theme
spread uniformly across all 20 topics has maximum entropy
(log\textsubscript{2}(20) = 4.32 bits). We report normalised entropy (0
to 1 scale) so that themes can be compared directly.

Second, we counted topic coverage: the number of discourse topics
containing at least one or at least five probes for each theme. This
directly measures how much of the discourse landscape contains each
concern. We also computed article coverage: the number and fraction of
unique source documents in the topics capturing 90\% of each theme's
probes.

Third, we computed a frame-to-topic compression ratio: the number of
unique survey response frames for each theme (from Lovato et al., 2024)
divided by the number of topics they occupy. A higher ratio indicates
greater opinion diversity being collapsed into fewer discourse regions.

\subsubsection{Reproducibility and use of large language
models}\label{reproducibility-and-use-of-large-language-models}

Analysis code, seed lists, query terms, probe templates, consensus UMAP
implementation details, and the full topic inventory are available at
https://github.com/AArtist1/When-Algorithms-Meet-Artists. Artist survey
data are available through Lovato et al.~(2024).

We used multiple LLMs across three pipeline stages: Likert anchor
generation for public probe retrieval, topic interpretation (described
above), and independent probe-theme validation (described below). At
each stage we employed multiple models and required human oversight to
mitigate single-model biases (Ziems et al., 2024). To validate the
public probe theme assignments, we classified the full 750-probe set
against the five concern themes using two independent large language
models, GPT-5.4-mini (OpenAI) and Claude Opus 4.6 (Anthropic), under a
multi-label prompt. For the four themes whose survey instrument carries
the theme concept directly (threat, utility, ownership, transparency,
N=600), mean multi-label recall was 0.742 (95\% CI 0.705, 0.775) for
GPT-5.4-mini and 0.733 (0.697, 0.767) for Claude Opus 4.6. Inter-LLM
agreement across all 3,750 probe-theme decisions was Cohen's \(\kappa\)
= 0.795 (95\% Wilson CI 0.773-0.816, n = 3,750, substantial agreement
per McHugh, 2012), with the two models producing identical multi-label
sets for 70.4\% of probes. Compensation was analysed separately because
its extraction showed systematically lower agreement. Only 27 of 150
compensation probes (18.0\%, 95\% Wilson CI 12.7-24.9\%) were confirmed
by both LLMs and then verified by human annotators. The rejected probes
converge on ownership (46\%) and utility (40-50\%) labels, directly
mirroring the frame-redirection pattern reported in the Results (see
Supplementary Information). We additionally validated probe construction
using 38 organic free-text responses from the Lovato et al.~survey,
finding that 84\% land in the same topic region as the corresponding
template probes (Supplementary Information). Machine-readable provenance
logs documenting AI-human collaboration during manuscript preparation
are available in the project repository, generated by an open-source
protocol for logging AI-assisted workflows (Author, 2026).

\subsection{Results}\label{results}

\subsubsection{Overview}\label{overview}

We constructed a semantic map of public AI-art discourse from 125
documents (1,736 text chunks) spanning 2013 to 2025, including news
articles, podcasts, panel discussions, academic papers, and legal
filings. Clustering identified 20 distinct topics organised into five
macro-thematic groups (Table 2; Figure S1). The largest groups,
Philosophy of Creativity (38.1\%) and Practice and Pedagogy (34.8\%),
together account for nearly three-quarters of the discourse, reflecting
the prominence of speculative and experiential narratives in public
visibility. Full topic descriptions are provided in our online
repository.

\textbf{Table 2: Macro-thematic groupings of the 20 public discourse
topics}

\begin{longtable}[]{@{}
  >{\raggedright\arraybackslash}p{(\linewidth - 4\tabcolsep) * \real{0.3919}}
  >{\raggedright\arraybackslash}p{(\linewidth - 4\tabcolsep) * \real{0.3514}}
  >{\raggedright\arraybackslash}p{(\linewidth - 4\tabcolsep) * \real{0.2297}}@{}}
\toprule\noalign{}
\begin{minipage}[b]{\linewidth}\raggedright
\textbf{Macro-theme}
\end{minipage} & \begin{minipage}[b]{\linewidth}\raggedright
\textbf{Share of corpus}
\end{minipage} & \begin{minipage}[b]{\linewidth}\raggedright
\textbf{Topics}
\end{minipage} \\
\midrule\noalign{}
\endhead
\bottomrule\noalign{}
\endlastfoot
Philosophy of Creativity & 38.1\% & 5 \\
Practice and Pedagogy & 34.8\% & 4 \\
Technical Genealogy & 14.7\% & 4 \\
Governance and Rights & 7.5\% & 3 \\
Institutions and Markets & 4.9\% & 4 \\
\end{longtable}

Into this semantic space, we projected 1,259 probes derived from survey
responses of 252 US-based practising artists, capturing 70 unique frames
across five concern dimensions: Threat, Utility, Ownership,
Transparency, and Compensation. We also projected 750 style-matched
public probes (short declarative sentences extracted from public
discourse) to control for stylistic differences between survey and media
formats.

\subsubsection{Artist concerns concentrate in a narrow discursive
region}\label{artist-concerns-concentrate-in-a-narrow-discursive-region}

\includegraphics[width=6.5in,height=5.04167in]{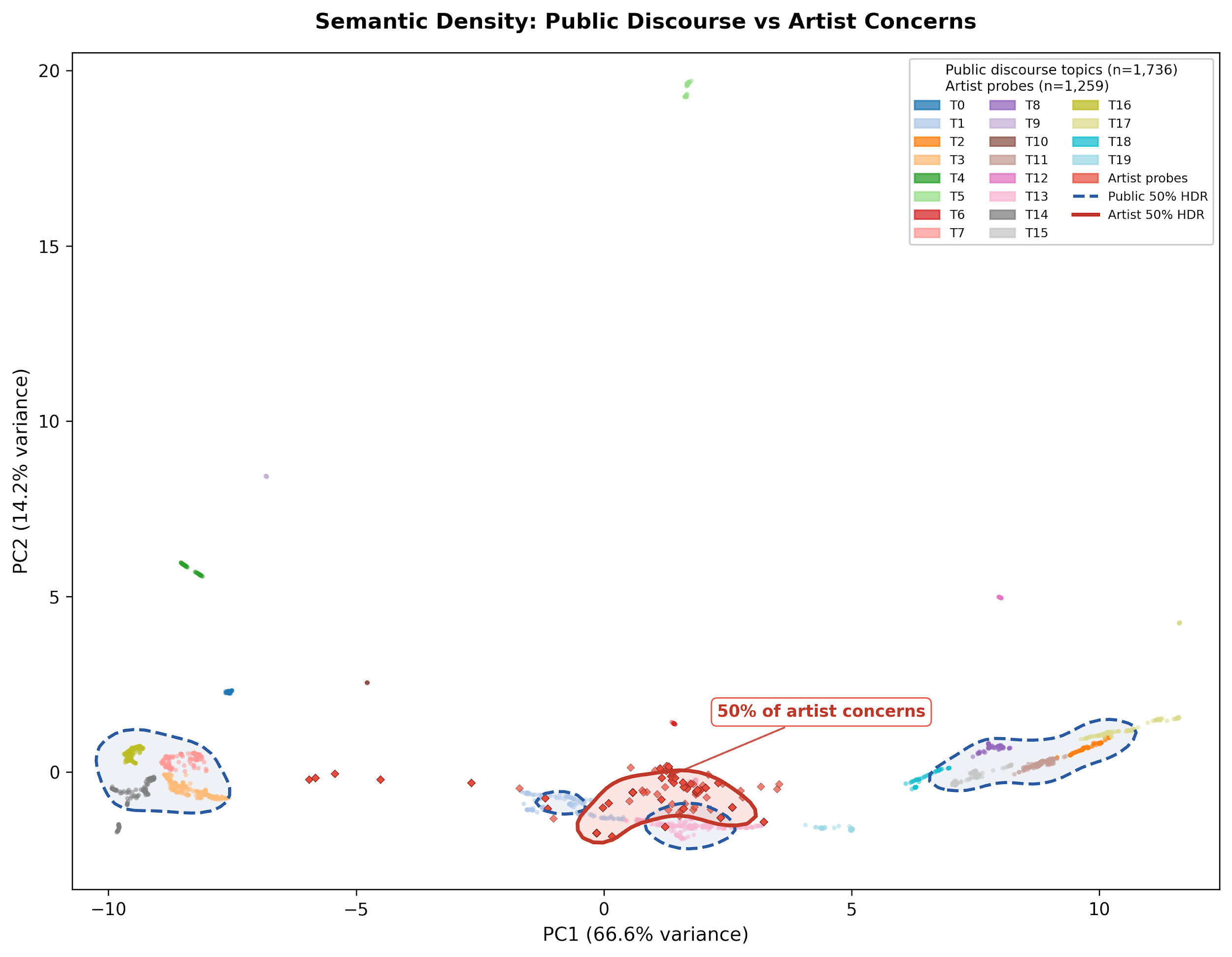}

\emph{\textbf{Figure 1.} Semantic density of public discourse and artist
concerns in shared embedding space (PCA projection of 5-dimensional
consensus UMAP coordinates, 80.8\% of variance captured in two
dimensions). Public discourse chunks are coloured by their KMeans topic
assignment (20 topics, tab20 palette). Artist probes are overlaid as red
diamonds. Bold contours enclose the 50\% highest-density region for each
distribution. The artist HDR (solid red) occupies approximately
one-third the area of the public discourse HDR (dashed blue),
visualising the semantic compression reported in the Results}.

When artist probes are projected into the public discourse semantic
space, a clear concentration pattern emerges (Figure 1). Two of the 20
public discourse topics absorb 94.9\% of all artist concerns (top-2
concentration permutation test against uniform-over-20-topics null:
observed = 0.948, n = 1,259, P \textless{} 0.0001, one-tailed) (Figure
2): ``AI Art Authenticity and Human Creativity'' (T13, 51.1\%, 643
probes) and ``AI as Creative Collaborator'' (T1, 43.8\%, 551 probes). A
further 5.0\% lands in two tail topics, ``Conversational Reflections on
Art Practice'' (T7, 4.4\%, 56 probes) and ``Personal Reflections on AI
and Art'' (T3, 0.6\%, 8 probes), so that four topics cover 99.9\% of
artist content in total (top-4 concentration permutation: observed =
0.999, P \textless{} 0.0001), with a fifth topic (``Generative Art
History and Pioneers'', T19) receiving one additional probe (0.08\%).
The remaining 15 topics contain zero artist perspective whatsoever, and
they comprise 58.2\% of public discourse volume. The regions where
artists are absent are dominated by institutional narratives (museum
exhibitions, NFT markets, funding models), technical genealogies
(DeepDream, GANs, diffusion models), and philosophical speculation about
machine creativity.

\includegraphics[width=6.5in,height=5.30556in]{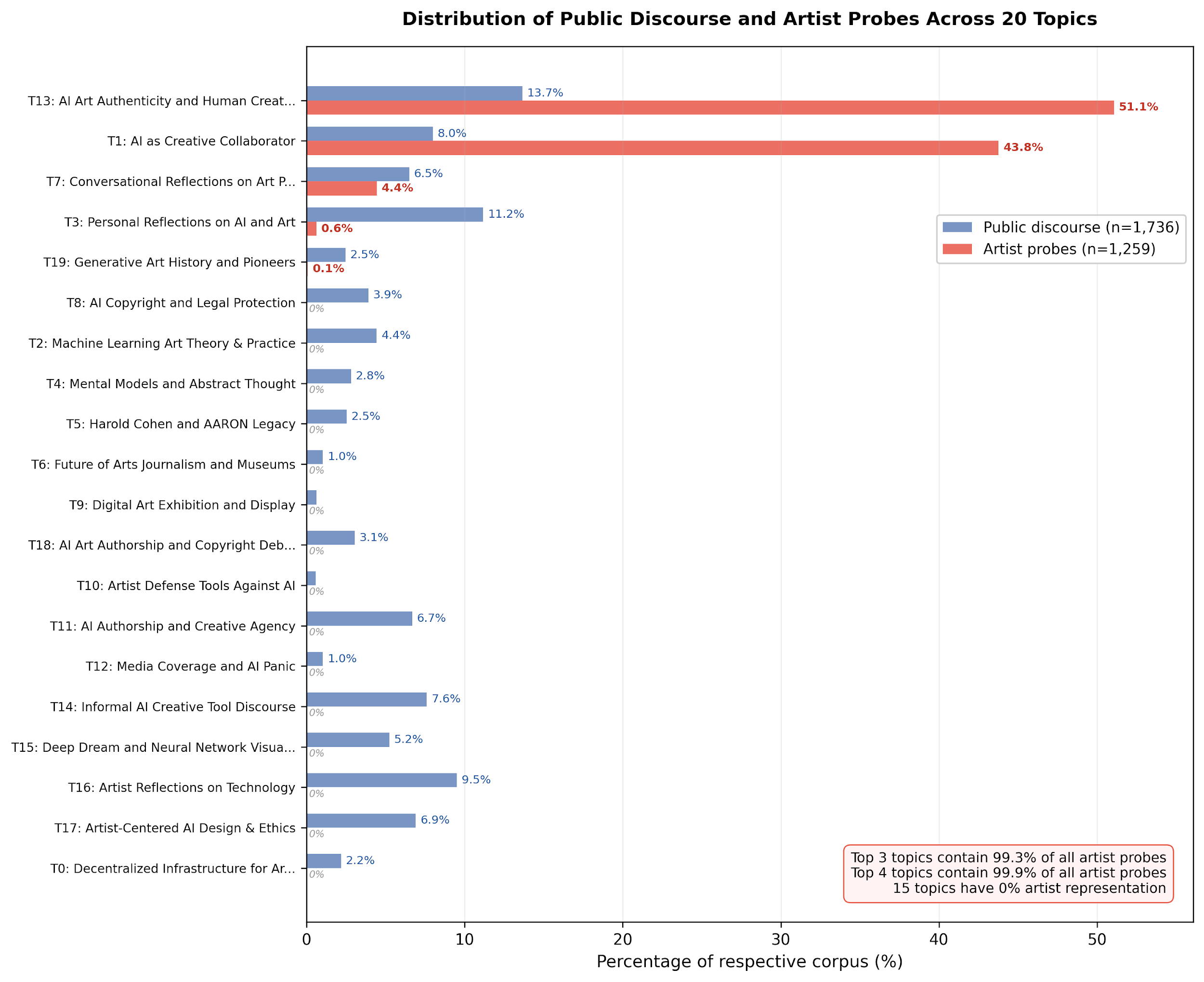}

\emph{\textbf{Figure 2.} Artist probe concentration across the 20 public
discourse topics. Two topics absorb 94.9\% of artist concerns on their
own (T13 ``AI Art Authenticity and Human Creativity'' at 51.1\% and T1
``AI as Creative Collaborator'' at 43.8\%), with two tail topics
carrying most of the remainder (T7 ``Conversational Reflections on Art
Practice'' at 4.4\% and T3 ``Personal Reflections on AI and Art'' at
0.6\%), bringing four-topic coverage to 99.9\%. 15 topics, comprising
58.2\% of public discourse volume, contain no artist perspective
whatsoever.}

This concentration is not an artifact of limited artist diversity. As
Table 1 illustrates, the 137 unique frame combinations in our sample
include positions the media binary cannot accommodate. The 55 artists
who view AI as both a threat and a useful tool would need to be split in
half to fit the ``Threat versus Tool'' categories that dominate public
coverage. Yet the entire range of positions, from the Pragmatic
Dual-Holder who would donate their work freely to the Protective
Advocate who demands revenue sharing, collapses into the same narrow
region of the public discourse space (Figure 1).

The compression is most acute as frame redirection. All 24 ownership
opinion frames map to a single discourse topic, and all three utility
positions (agree, disagree, unsure) map to that same topic. Both
dimensions land entirely in T13, ``AI Art Authenticity and Human
Creativity.'' Compensation spreads across 5 topics, but 96.4\% of
compensation probes still concentrate in three of them.

Artist frames concentrate in a small subset of public discourse topics,
with the 70 frames collapsing into just two primary destinations (a 35:1
ratio, with a longer tail reaching four topics that together carry
99.9\% of probes). The discursive space to represent these concerns
exists. Artist voices are simply not placed within it.

\subsubsection{Marginalisation is semantic, not
stylistic}\label{marginalisation-is-semantic-not-stylistic}

A potential concern is that the observed separation reflects stylistic
mismatch (survey responses are short declaratives, while public
discourse includes long-form articles and transcribed speech) rather
than substantive semantic differences. We addressed this by comparing
artist probes to style-matched public probes: short declarative
sentences extracted from public documents that match the syntactic form
of survey responses.

If the artist-public separation were primarily stylistic, it should
collapse when both corpora share the same format. It does not.

\textbf{Distributional evidence.} Style controls reduce Cramér's V from
0.740 (\(\chi^{2} = 1639.5\), df = 19, n = 2,995, P \textless{} 0.0001,
two-sided) to 0.734 (\(\chi^{2} = 1081.1\), df = 12, n = 2,009, P
\textless{} 0.0001), a 0.8\% reduction in effect size. Jensen-Shannon
divergence falls from 0.364 (10,000-permutation test, P \textless{}
0.0001, one-tailed) to 0.308 (P \textless{} 0.0001, one-tailed), a
15.4\% reduction. Format differences between survey statements and media
text account for a modest portion of the raw distributional divergence.
Cramér's V of 0.734 remains well above the threshold for large effects
(V \textgreater{} 0.5), and the corpora remain geometrically separated
in the consensus space.

\textbf{Geometric evidence.} In the consensus semantic space, artist
probes are separated from raw public discourse by a centroid distance of
2.003 units (10,000-permutation test against label-shuffle null, P
\textless{} 0.0001, one-tailed) and from the style-matched public probes
by 3.754 units (P \textless{} 0.0001, one-tailed). Instead of collapsing
it, the style control widens the geometric separation. 83.4\% of public
probes have other public probes as their nearest neighbours rather than
mixing with artist probes, and 99.9\% of artist probes have other artist
probes as nearest neighbours when compared to raw public discourse. This
indicates that artists occupy a geometrically distinct region of the
semantic manifold. Their probes are not merely expressing the same ideas
differently. Compression patterns are also invariant to artists'
self-reported AI exposure (use of AI art models and familiarity with AI
model categories; see Supplementary Information).

\textbf{Interpretation.} Format differences contribute to the observed
divergence but do not account for it. A modest fraction of the raw JSD
is attributable to stylistic mismatch between survey and media formats.
The Cramér's V is essentially unmoved, indicating that the categorical
association between corpus and topic distribution is not a function of
format. Even when public discourse is segmented into short declarative
sentences matching the survey format, it does not converge with artist
concerns. Institutional and technical narratives dominate public
discourse even at the sentence level, while artist concerns about
consent, compensation, and creative rights remain peripheral.

Distributional and geometric divergence persist after style controls.
The divergence is predominantly semantic, with format differences
accounting for a modest portion of the raw JSD (15.4\% reduction).

\subsubsection{Compression operates at multiple
levels}\label{compression-operates-at-multiple-levels}

The 1,259 artist probes do not distribute uniformly across the 20
discourse topics. They concentrate in just 5 topics, leaving 15 with no
artist perspective at all (Figure 2). To understand the structure of
this compression, we examined which topics absorb artist voices, what
those topics are labelled, and how artist positions within each concern
dimension distribute across the landscape. The pattern that emerges is
not a single compression mechanism but a layered set of operations.
These include topical exclusion, frame redirection, binary
simplification, and voice collapse.

\textbf{Table 3: Compression metrics by artist concern theme}

\begin{longtable}[]{@{}
  >{\raggedright\arraybackslash}p{(\linewidth - 12\tabcolsep) * \real{0.1181}}
  >{\raggedright\arraybackslash}p{(\linewidth - 12\tabcolsep) * \real{0.1024}}
  >{\raggedright\arraybackslash}p{(\linewidth - 12\tabcolsep) * \real{0.1811}}
  >{\raggedright\arraybackslash}p{(\linewidth - 12\tabcolsep) * \real{0.1654}}
  >{\raggedright\arraybackslash}p{(\linewidth - 12\tabcolsep) * \real{0.1890}}
  >{\raggedright\arraybackslash}p{(\linewidth - 12\tabcolsep) * \real{0.1496}}
  >{\raggedright\arraybackslash}p{(\linewidth - 12\tabcolsep) * \real{0.0787}}@{}}
\toprule\noalign{}
\begin{minipage}[b]{\linewidth}\raggedright
\textbf{Theme}
\end{minipage} & \begin{minipage}[b]{\linewidth}\raggedright
\textbf{Frames}
\end{minipage} & \begin{minipage}[b]{\linewidth}\raggedright
\textbf{Artist Consensus}
\end{minipage} & \begin{minipage}[b]{\linewidth}\raggedright
\textbf{Entropy (norm)}
\end{minipage} & \begin{minipage}[b]{\linewidth}\raggedright
\textbf{Topics (any / 5+)}
\end{minipage} & \begin{minipage}[b]{\linewidth}\raggedright
\textbf{Articles (\%)}
\end{minipage} & \begin{minipage}[b]{\linewidth}\raggedright
\textbf{FCR}
\end{minipage} \\
\midrule\noalign{}
\endhead
\bottomrule\noalign{}
\endlastfoot
Ownership & 24 & 17.9\% top response & 0.000 & 1 / 1 & 34 (27\%) &
24.0 \\
Utility & 3 & 46.0\% agree & 0.000 & 1 / 1 & 34 (27\%) & 3.0 \\
Transparency & 3 & 81.0\% agree & 0.163 & 2 / 2 & 46 (37\%) & 1.5 \\
Threat & 3 & 60.6\% agree & 0.176 & 2 / 2 & 46 (37\%) & 1.5 \\
Compensation & 37 & 18.7\% top response & 0.351 & 5 / 4 & 52 (42\%) &
7.4 \\
\end{longtable}

\emph{Entropy is normalised against the maximum possible
(log}\textsubscript{2} \emph{of topic count). Artist Consensus reports
the proportion holding the most common position within each dimension.
FCR = frame-to-topic compression ratio (unique frames divided by the
number of occupied topics, where an occupied topic contains at least one
probe from that theme). Ownership and utility achieve the most extreme
compression (Figure 3A-C), with every probe from each dimension mapping
to a single discourse topic.}

Of the 20 public discourse topics, only 5 contain any artist probes. The
topics that absorb artist voices are: AI Art Authenticity and Human
Creativity (T13: 643 probes, 51.1\%), AI as Creative Collaborator (T1:
551, 43.8\%), Conversational Reflections on Art Practice (T7: 56,
4.4\%), Personal Reflections on AI and Art (T3: 8, 0.6\%), and
Generative Art History and Pioneers (T19: 1, 0.08\%). The remaining 15
topics, comprising 58.2\% of public discourse volume, contain zero
artist perspectives.

A closer reading of the 15 empty topics reveals a systematic absence.
Three of these topics received labels the LLM-and-human panel read as
explicitly about artist rights: AI Copyright and Legal Protection (T8,
68 chunks), Artist Defense Tools Against AI (T10, 10 chunks), and AI Art
Authorship and Copyright Debates (T18, 53 chunks). Together these
contain 131 public discourse chunks that the annotators interpreted as
concerning copyright disputes, legal protection, and defensive
technologies. Three further zero-artist topics received labels the
annotator panel read as artist-themed: AI Authorship and Creative Agency
(T11, 116 chunks), Artist Reflections on Technology (T16, 165 chunks),
and Artist-Centered AI Design and Ethics (T17, 120 chunks). Across these
six topics, 532 chunks of public discourse fall in regions whose panel
reading foregrounds artists or their rights. Public discourse has built
substantial conversational infrastructure for talking about artists and
artist legal protection without using language that maps to how artists
themselves articulate those concerns. Artists are the subject of these
conversations but not their participants.

Artist concerns about ownership and utility do not vanish from the
discourse map. They are redirected into a single topic. T13, which the
annotator panel labelled "AI Art Authenticity and Human Creativity,"
absorbs all 252 ownership probes (100\%) and all 252 utility probes
(100\%), plus a portion of transparency, compensation, and threat (643
artist probes in total, 51.1\% of all artist content). The panel-derived
label, produced by independent reading of T13\textquotesingle s most
representative chunks by four LLMs and two human coders, points to a
philosophical discussion of authenticity, yet the cluster also contains
every concrete property claim and every utility position artists make.
The 24 unique ownership frames, spanning fundamentally different views
on creative property, all collapse into a discourse region whose
dominant readable theme is aesthetic authenticity rather than rights or
labour.

\includegraphics[width=6.5in,height=6.79167in]{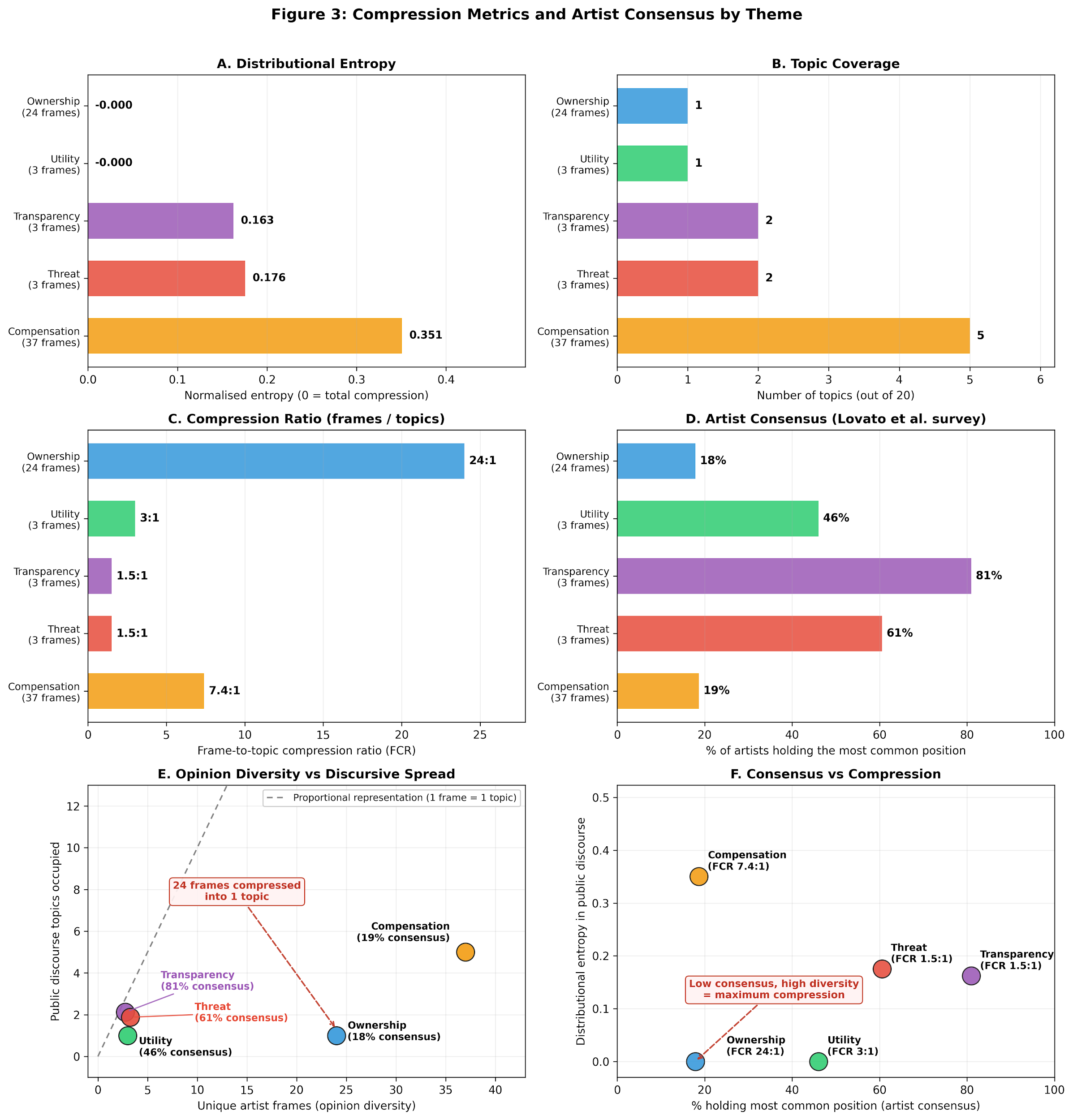}

\emph{\textbf{Figure 3.} Compression metrics by artist concern theme.
(A) Distributional entropy by theme. Ownership and utility show total
compression (entropy = 0.000, n = 252, P \textless{} 0.0001, one-tailed
permutation vs uniform null), with all 252 probes from each dimension
landing in a single topic. Transparency (0.163, N = 252, P \textless{}
0.0001) and threat (0.176, n = 251, P \textless{} 0.0001) show near-zero
entropy in two-topic splits. (B) Topic coverage: number of discourse
topics containing probes for each theme, out of 20 total. (C)
Frame-to-topic compression ratio: ownership has the highest ratio (24:1)
when 24 distinct opinion frames collapse into one topic. (D) Artist
consensus from the Lovato et al.~survey: transparency has the highest
consensus (81\% agree on mandatory disclosure) yet near-zero entropy in
public discourse. (E) Opinion diversity (number of unique frames per
theme) plotted against the number of public discourse topics each theme
occupies, with the proportional representation reference line. (F)
Artist consensus plotted against distributional entropy in public
discourse, showing that consensus does not protect against compression
(Figure 3F): transparency carries the highest artist consensus yet still
collapses to near-zero entropy in public discourse.}

This is frame compression at its most severe (Figure 3C). Akrich (1992)
calls this mismatch between inscribed and actual users the description
gap. In discourse it produces the effect that ownership claims are
processed as contributions to a conversation their speakers did not
enter: the philosophical-aesthetics conversation T13's label inscribes,
rather than the property-rights conversation the artist is making.

The threat dimension reveals the binary structure that prior media
studies have documented. The 251 threat probes split cleanly into two
topics:

\begin{itemize}
\item
  ``I agree that AI art models are a threat to art workers'' (152
  probes) and ``I am unsure'' (44) \(\rightarrow\) AI as Creative
  Collaborator (T1: 196 probes total, 78.1\%)
\item
  ``I disagree that AI art models are a threat to art workers'' (55
  probes, 21.9\%) \(\rightarrow\) AI Art Authenticity and Human
  Creativity (T13)
\end{itemize}

The split is clean, but the destinations invert any straightforward
reading. The position ``AI is a threat to art workers'' lands in a topic
literally labelled ``AI as Creative Collaborator,'' a positive framing
of AI. The position ``AI is not a threat'' lands in a topic about
authenticity philosophy. Neither destination preserves the labour
argument the threat statement makes. This mirrors the AI good versus AI
bad binary that media analyses have repeatedly identified in technology
coverage (Entman, 1993; Bøgh, 2025; Cheung, 2025; Saurwein et al.,
2025), now visible at the structural level of the semantic map itself.

Topic T13 absorbs all three utility positions in their entirety: 116
probes that affirm AI is a positive development for art, 69 that affirm
it is not a positive development, and 67 that express uncertainty. These
mutually opposite positions land at the same coordinates not because of
substantive content but because they share first-person, practice-voice
phrasing about AI and art. All three positions are expressed in
first-person, practice-voice statements about AI and art, and the
semantic map groups them on that basis. Three substantively opposite
positions therefore land at the same coordinates.

The same voice-collapse logic operates inside T1. That topic pulls
together 204 probes demanding mandatory disclosure (transparency), 151
compensation positions, 152 probes affirming AI is a threat, and 44
probes that are unsure. These are heterogeneous substantive positions
with one thing in common, a register of concerned, first-person artist
commentary on AI's impact. The topic groups them by tone, not by claim.
Figure S2 makes this voice collapse visible across all five concern
dimensions: the 95\% convex hulls for threat, utility, ownership, and
transparency occupy a narrow overlapping region in the consensus space.
Only compensation shows meaningful spatial spread.

Out of 252 transparency probes, 204 (81\%, the ``agree'' position) map
to T1, AI as Creative Collaborator. The remaining 48 (the ``unsure'' and
``disagree'' positions, 19\%) map to T13, AI Art Authenticity and Human
Creativity. Normalised entropy is 0.163 (n = 252, P \textless{} 0.0001,
one-tailed permutation vs uniform null). The discourse semantic space
barely differentiates pro-disclosure artists from those who oppose
disclosure. A theme with the highest stakeholder consensus (81\%
agreement on mandatory disclosure) is also one of the most spatially
compressed. Nearly every probe lands in two topics whose labels
reference creative collaboration and aesthetic authenticity, neither of
which captures the regulatory demand at the heart of the transparency
claim. Consensus does not protect against compression. It may make
compression easier, since the discourse can collapse a near-unanimous
position into a single representative slot.

Figure 3 shows these patterns across all five themes. Ownership and
utility achieve total compression at zero entropy because every probe
from each dimension maps to one topic, and that topic (T13) is
pre-framed as a debate about authenticity rather than property or
labour. Transparency near-totally compresses, with 81\% in one topic and
19\% in a second, neither of which preserves the regulatory demand.
Threat shows the canonical media binary, splitting cleanly into two
topics whose labels invert any plain reading of the underlying
positions. Compensation spreads across five topics (entropy 0.351)
because its 37 distinct positions span enough rhetorical variation that
the semantic map can detect, but 96.4\% of compensation probes still
concentrate in three topics.

Compression in public discourse operates not as a single mechanism but
as a layered set of operations: topical exclusion (15 of 20 topics
contain no artist voice), frame redirection (ownership and utility
concerns get rerouted into a single topic pre-framed as a debate about
artistic authenticity), binary simplification (threat resolves into two
topics whose labels invert the substantive positions), and voice
collapse (opposite positions on the same question group together because
they share first-person phrasing rather than substantive content). The
panel-derived labels assigned to the topics that exist in public
discourse but receive no artist probes (AI Copyright and Legal
Protection, Artist Defense Tools Against AI, AI Art Authorship and
Copyright Debates, AI Authorship and Creative Agency, Artist Reflections
on Technology, Artist-Centered AI Design and Ethics) indicate that the
LLM-and-human annotators read these clusters as carrying conversations
about artist rights and creative agency. That is, the absence of artist
probes is not because no readable infrastructure for such conversations
exists in the corpus. It exists. Artists are simply not present in it.

\subsection{Discussion}\label{discussion}

\subsubsection{Overview}\label{overview-1}

Our results show that public discourse on AI and art systematically
compresses stakeholder complexity into a small subset of the available
framings. While artists articulate a high-dimensional landscape spanning
nuanced positions on threat, utility, ownership, transparency, and
compensation, public discourse collapses this variance into a narrow
region of the semantic map (Figure 1). Across distributional, geometric,
and stratified analyses, artist stakeholder frames are not
proportionally represented in the public discourse topic landscape, even
after accounting for stylistic form.

The compression we document has a specific character that distinguishes
it from the patterns reported in prior work. Studies of AI discourse
across Western media (Zai et al., 2025), Chinese education coverage
(Huang and Gadavanij, 2025), and ChatGPT social media discussions (Cohen
et al., 2026) have all found that affected stakeholders receive less
discursive standing than institutional actors. We extend these accounts
by measuring the structure of compression and identifying which
stakeholder concerns are most affected: underrepresentation operates
through topical absence, frame reclassification, binary splitting, and
voice-level grouping operating together (Figure 3). This selective
pattern, which we characterise as marginalisation without explicit
exclusion, means the public discourse environment is not uniformly
hostile to artist voices. It is structurally hostile to the voices that
matter most for policy.

The magnitude of this compression is worth emphasising. The 70 frames in
our artist sample span positions that are not merely different but often
mutually incompatible: artists who would license their work through
opt-in registries sit alongside those who reject all property
frameworks. Those who demand retroactive compensation coexist with those
who would donate freely to a commons (Lovato et al., 2024; Kawakami and
Venkatagiri, 2024). Yet this internal political complexity maps onto a
small fraction of the public discourse space.

\subsubsection{The phenomenon of semantic
compression}\label{the-phenomenon-of-semantic-compression}

The public discourse landscape on AI art appears pluralistic at first.
Our semantic map contains 20 topics spanning technical genealogies,
institutional and market narratives, governance and law, philosophy of
creativity, and tool practice. This breadth matches recent topic-mapping
work showing that public debates about image-generative AI contain
multiple distinct interpretive topics rather than a single linear
narrative (Saeedi and Taleghani, 2025; Banks and Li, 2025).

We are interested in identifying which topics become visible across the
semantic space. Entman (1993) identifies four functions of framing:
problem definition, causal interpretation, moral evaluation, and
treatment recommendation. Compression cuts across all of these
functions. Public discourse defines the problem of AI art as spectacle
and disruption and reads causation through technological determinism.
Moral stakes are evaluated through binary threat-versus-opportunity
frames, and the treatments recommended reflect institutional, rather
than practitioner, priorities.

This framing distortion is well-documented across AI domains. In AI-art
media coverage, nuanced labour concerns are reduced to ``AI as Threat''
versus ``AI as Tool'' framings (Bøgh, 2025). In longitudinal coverage of
virtual assistants, discourse transitioned from personal-impact to
societal-concern frames but consistently privileged institutional and
technological narratives over user experience (Wald et al., 2026). Our
results extend this logic. Stakeholder-derived frames occupy a
comparatively narrow region of the public topic space.

The public agenda has set the issue of AI art without setting the
attributes that practitioners find most salient, a pattern consistent
with second-level agenda-setting failure.

Public discourse does not allocate proportional representational space
to this range of stakeholder positions. The Pragmatic Dual-Holder and
the Protective Advocate from Table 1 both land in the same topic,
despite holding fundamentally different positions on utility and
compensation. Public discourse registers them both as ``artist
backlash.'' The distinction between an artist who would donate their
work freely and an artist who demands revenue sharing is erased.
Further, all 24 ownership opinion frames collapse into a single topic.
The public agenda has stripped the issues of AI from their internal
political logic, reducing 137 distinct combinations of stakeholder
positions to broad tropes of ``resistance'' or ``democratisation.''
Campos Valverde and Kaye (2026) document a parallel dynamic in the music
industry, where ethnographic research at trade conferences reveals three
competing positions on generative AI (protectionist, liberalising, and
conciliatory) that are compressed in public-facing industry discourse.
Their finding that institutional framing channels and redirects creator
concerns parallels the semantic compression we identify in AI-art media
coverage. Our own sample reinforces this point. It includes writers,
musicians, designers, and craftspeople alongside visual practitioners,
yet every concern dimension is absorbed into a public discourse whose
dominant cases are visual (Midjourney, Stable Diffusion, Andersen v.
Stability AI). The compression operates not only within a creative
modality but across modalities, with multi-modal stakeholder concerns
funnelled through a predominantly image-AI public frame.

\subsubsection{Mechanisms of
compression}\label{mechanisms-of-compression}

The compression we document is not a single process but a layered set of
operations. Each operates on a different aspect of artist speech, and
each leaves a distinct signature in the projected geometry. We treat
them in turn because their policy implications differ, and because prior
literature has typically named only one or two in isolation, without
tracing how they combine. Together they illustrate what Bowker and Star
(1999) describe in terms of \emph{residual categories}, where a
classificatory system renders particular positions invisible by
absorbing them into categories that cannot recognise them as political
claims.

\paragraph{Topical exclusion}\label{topical-exclusion}

Fifteen of the twenty discourse topics contain no artist probe at all
(Figure 2). These are not minor or peripheral topics. They include three
topics explicitly about artist legal protection (AI Copyright and Legal
Protection, Artist Defense Tools Against AI, AI Art Authorship and
Copyright Debates) and three further topics with artist-themed labels
(AI Authorship and Creative Agency, Artist Reflections on Technology,
Artist-Centered AI Design and Ethics). Together these six topics contain
532 chunks of public discourse, fully 30.6\% of the corpus, that talk
about artists and artist concerns without ever using language that maps
to how artists themselves articulate those concerns. The conversation
has built rooms for these topics. Artists are not in any of them. This
is where the conversation about rights proceeds without the people whose
rights are at stake. The topic exists, the topic is populated with
public-discourse chunks, and yet the voices that would most directly
articulate what rights mean in this domain are located elsewhere in the
semantic geometry.

The absent rooms are populated instead by voices orthogonal to artist
experience: curators, executives, and platform operators discussing
funding models and NFT economies, technical genealogies rehearsing
DeepDream (Mordvintsev et al., 2015), GANs (Goodfellow et al., 2014),
and diffusion milestones as spectacle, and doctrinal-philosophical legal
framings that exclude practitioner testimony. The same institutional
asymmetry appears in AI-education discourse (Huang and Gadavanij, 2025)
and in automation coverage, where industry stakeholders absorb 45\% of
responsibility attributions while users and civil society each receive
7\% (Saurwein et al., 2025). Bishop (2025) documents the corollary
inside creative labour: "influencer creep" migrating platform
optimisation logics into artistic practice. These lived realities of
creative workers navigating algorithmic systems are absent from the
institutional narratives dominating public AI-art discourse.

This is the strongest empirical signature of epistemic marginalisation
without explicit exclusion, and it instantiates at the level of public
discourse what Widder (2024) identifies inside AI ethics labour itself.
The topics exist, the labels are artist-facing, and the epistemic
authority over what those topics contain rests elsewhere. The discourse
environment is not silent on artist rights. It is verbose on artist
rights and silent only with respect to artist voices.

\paragraph{Frame redirection}\label{frame-redirection}

Where do ownership and utility actually go? Both dimensions achieve
total compression at zero entropy. Every one of the 252 ownership probes
and every one of the 252 utility probes maps to a single topic, T13,
labelled ``AI Art Authenticity and Human Creativity.'' The label points
to a philosophical debate about whether AI-generated work qualifies as
art. The contents include 24 distinct opinion frames on who should own
AI-generated artwork and three opposite positions on whether AI is
useful for art practice. An artist who states ``I disagree that
AI-generated artwork in my style should be considered the property of
the AI user'' is making a property claim. The semantic map routes that
claim into a discourse region pre-framed as an aesthetic question. Its
closest discursive neighbour is not ``artist demands ownership'' but
``is AI art real art?'' The topic label inscribes an imagined
interlocutor, a philosopher of aesthetics asking about the ontological
status of AI-generated images, whose concerns do not match those of the
artists whose speech actually lands there. Akrich (1992) calls this
mismatch between inscribed and actual users the \emph{de-scription} gap.
In discourse it produces the effect that claims about property and
labour are processed as contributions to a conversation their speakers
did not enter.

Frame redirection is more consequential than topical exclusion. An
excluded topic at least preserves the absence as visible. A reader
scanning the topic inventory can see that no artist probe appears in the
AI Copyright topic. A redirected frame, by contrast, hides the
substantive claim by reclassifying it. A policymaker reading the AI Art
Authenticity topic will encounter discussion of authorship, originality,
and aesthetic value. They will not see that this is also where every
ownership claim and every utility position gets routed. The frame
preserves the words but loses the demand.

\paragraph{Binary simplification}\label{binary-simplification}

Media analyses of AI debates have repeatedly documented binary framing:
opportunity versus crisis, tool versus threat, hype versus harm (Entman,
1993; Bøgh, 2025; Cheung, 2025; Saurwein et al., 2025). Saurwein et
al.~(2025) note that responsibility attributions in automation coverage
resolve into oppositional networks rather than distributed accounts,
with industry stakeholders absorbing 45\% of attributions. Our data
shows this binary operating not as a rhetorical choice in any individual
article but as a structural property of the semantic map itself.

The 251 threat probes split into exactly two topics. The 152 probes
affirming "AI is a threat to art workers" and the 44 probes expressing
uncertainty both land in T1, which the annotator panel labelled "AI as
Creative Collaborator." The 55 probes that disagree with the threat
statement land in T13, the topic the panel read as concerned with
authenticity philosophy. The split is clean, but the panel readings of
the destination clusters invert the substantive content of the artist
statements landing there. The position "AI is a threat" arrives in a
cluster whose dominant readable theme, on the panel reading, is creative
collaboration. The position "AI is not a threat" arrives in a cluster
whose readable theme, on the panel reading, is aesthetic authenticity.
Neither panel reading preserves the labour argument the threat statement
makes. The discourse offers no semantic space for a substantive claim
that AI does or does not constitute a threat to creative workers, only a
binary between concerned-artist phrasing and aesthetic-philosophical
phrasing. Małecki et al.~(2025) provide complementary experimental
evidence. Public perception of AI art is overwhelmingly shaped by the
binary question of whether a work is ``AI-generated'' or ``human-made,''
with attribution alone reducing aesthetic appreciation. Our findings
extend their result. The ``AI versus human'' binary is not only a
cognitive default at the level of perception. It is also a structural
default at the level of how public discourse organises itself around
artist speech.

The same logic operates inside T1. That topic pulls together 204
transparency-agree probes, 151 compensation positions, 152 threat-agree
probes, and 44 threat-unsure probes: 551 substantively heterogeneous
statements with one thing in common, a register of concerned artist
commentary. The topic is not a topic in the conventional sense. It is a
voice signature.

\paragraph{Voice collapse}\label{voice-collapse}

Voice collapse has a specific governance consequence. A policymaker
reading either topic would conclude that the artists in it share a
position. They do not share a position. They share a way of speaking.
The discourse cannot distinguish artists who welcome AI from artists who
reject it when both speak in the same personal register about their
craft. This is the level at which the compression becomes hardest to
detect, because the topic appears coherent until one reads its contents.

These operations compound. Artists are removed from topics that bear
their own name, and their property and labour claims are rerouted into
aesthetic-philosophy discourse, stripped of substantive demand. The
threat dimension, where it surfaces, splits into oppositional pairs
whose labels invert the positions they contain, while inside the topics
that do carry artist voice, opposite positions on the same question
collapse into a shared practice-register and appear coherent only until
the contents are read. Each of these patterns is independently
documented in the prior media-framing literature. Their combination is
what the present analysis adds. The same artist concern can be absent
from one topic, redirected into another, binarised inside a third, and
voice-collapsed inside a fourth. The compression is not a single signal
but a system. Policymakers who rely on public discourse as a proxy for
stakeholder priorities will miss the internal political complexity of
the governance demands that should inform regulation, and they will miss
it through several distinct mechanisms operating simultaneously.

\subsubsection{Algorithmic
agenda-setting}\label{algorithmic-agenda-setting}

Agenda-setting in the contemporary discourse environment is increasingly
algorithmic, shaped by recommender systems and platform ranking
alongside editorial institutions (Sichach, 2024; Pane, 2025). Brause et
al.~(2025) extend Jasanoff's (2015) concept of sociotechnical
imaginaries to public communication, arguing that whose imagination of
AI becomes authoritative depends on the \emph{speaker} dimension of the
imaginary. Our compression findings offer an empirical audit of that
speaker dimension. In the AI-art imaginary, artists are objects of
discussion, not its authoritative speakers. The empirical pattern is
consistent with this account. Topic salience in our corpus is driven by
a small number of high-visibility events (DeepDream in 2015, the Belamy
auction in 2018, DALL-E and diffusion-model releases after 2021), events
that generate discourse about capability and spectacle rather than
labour impact, and the topics that do carry governance-related artist
probes appear in only 27-37\% of source articles, far below the
corpus-wide baseline.

\subsubsection{Coalition mobilisation accentuates the
stakes}\label{coalition-mobilisation-accentuates-the-stakes}

Recent organising by creator coalitions confirms that transparency,
consent, and compensation are not fringe concerns but actively contested
governance claims. In the United Kingdom, over 40 creative industry
organisations, including the Association of Illustrators, the Society of
Authors, the Publishers Association, and the Association of
Photographers, have mobilised through the Creative Rights in AI
Coalition since December 2024 against unauthorised use of copyrighted
works in AI training (Creative Rights in AI Coalition, 2024). The US
Creators Coalition on AI, launched in December 2025, pursues similar
objectives (Creators Coalition on AI, 2025). The \emph{Andersen v.
Stability AI} class action demonstrates that artist grievances have
reached federal litigation, with plaintiffs alleging that generative
models trained without consent constitute mass infringement (Andersen et
al.~v. Stability AI, 2023).

These developments strengthen the normative and policy relevance of
auditing representational patterns. Artist concerns are not hypothetical
or speculative. They are the subject of organised mobilisation,
legislative testimony, and active litigation. The gap between what
artists articulate and what public discourse makes salient is not simply
an academic observation. It is a governance failure with material
consequences.

\subsubsection{Implications for policy and
governance}\label{implications-for-policy-and-governance}

The semantic compression we document poses a direct risk to AI
governance. If policymakers rely on public sentiment analysis, media
monitoring, or ``public comment'' solicitation to gauge the impact of
generative AI on creative labour, they will receive a distorted signal.
They will see generic debates about ``AI and art'' but will miss the
distinctions that matter for regulation: between artists who want
opt-out mechanisms versus licensing arrangements, between those who
would donate freely versus those demanding ongoing compensation, between
those who accept AI as a tool versus those who view it as an existential
threat, and between those who prioritise transparency versus ownership.
Policymakers reading public discourse would not know these distinctions
exist. As Ferrari et al.~(2025) argue, effective governance of
generative AI requires structural conditions of observability,
inspectability, and modifiability, conditions that depend on stakeholder
input being adequately represented in the deliberative processes shaping
regulation. Our findings suggest that these conditions are not met when
public discourse compresses the very concerns that governance frameworks
must address. The risk is performative in MacKenzie's (2006) sense.
Governance instruments calibrated against a compressed discursive signal
do not merely fail to detect the marginalisation. They help reproduce it
by treating that compressed signal as an adequate representation of
stakeholder concerns.

Calls for stakeholder participation in AI governance are easily
satisfied by procedures that do not change which concerns become
consequential (Suresh et al., 2024). Our methodology (projecting
stakeholder frames into public discourse space and computing compression
metrics) offers a measurable check on whose positions achieve discursive
standing. Instead of assuming public discourse reflects stakeholder
priorities, we provide measurable indicators of mismatch. This approach
could be applied to other technology-affected stakeholder communities
(gig workers, content moderators, data subjects, data labourers) to
assess representational equity in the discourses that shape platform and
AI governance. In each case, concerns are liable to be filtered through
institutional and technical framings before reaching governance debates,
and a comparable compression dynamic may be observable. Operationalised
this way, semantic-compression auditing extends the data-feminist
principle that representational infrastructures should be examined for
whose knowledge they render visible and whose they obscure (D'Ignazio
and Klein, 2020), while acknowledging the methodological tension that
distributional measurement aggregates the very voices it seeks to
elevate.

\subsubsection{Limitations}\label{limitations}

We acknowledge several limitations:

\textbf{Methodological constraints:} Artist probes are structured
responses constrained by survey framing (Lovato et al., 2024), not
organic discourse. The public corpus is shaped by Google Search
retrieval, which reflects existing visibility hierarchies. Sources that
already rank highly are overrepresented, while artist-centric
publications, community forums, and non-English sources are likely
undersampled. Conversely, our exclusion of social media platforms
(Reddit, Twitter/X), where artists are often vocal, may have removed
discursive spaces where artist concerns achieve greater salience. Future
work should compare search-ranked, editorialised, and social-media-first
corpora to assess whether compression varies by platform type. Our
reliance on LLMs for anchor generation, validation filtering, and topic
interpretation introduces a dependency on specific model versions and
their training distributions. We mitigated this by using multiple models
and requiring human-LLM consensus, but cannot rule out that the models'
training data shape which topics and frames are detected. Our use of a
single embedding model (e5-large-v2) may introduce representational
biases. These constraints are not failings to be apologised for so much
as a \emph{located accountability} in Suchman's (2002) sense. Any
measurement of representational compression is itself produced from a
particular technical and epistemic position, with a partial view of the
discourse it characterises. Widder (2024) makes the related point that
quantified AI ethics work secures institutional legitimacy precisely
because it adopts the epistemic register of the systems it critiques. We
accept this tension: a distributional measurement of artist
marginalisation is offered here as a complement to, not a substitute
for, located, embodied, and artist-led accounts.

\textbf{Inferential constraints:} We document patterns consistent with
epistemic marginalisation but do not directly demonstrate causal
mechanisms. Coalition, agenda-setting, and algorithmic amplification
claims remain interpretive without evidence from actor-networks or
platform traces.

\textbf{Scope constraints:} Our analytic sample spans artists across
multiple creative practices, predominately painting (43\%), photography
(19\%), writing (11\%), and then design, music, digital, illustration,
sculpture, drawing, craft, and tattoo capturing the rest. It also
contains a public discourse corpus that captures visual (image), music,
video, writing, and film generation. Two scope notes follow. First, our
analysis treats artists as a single stakeholder group, without
disaggregating by practice modality. Differential representation across
modalities is an open empirical question that the Lovato et al.~dataset
can support. Second, the method of projecting stakeholder survey
responses into a public discourse semantic space is domain-general in
principle, but specific parameters (embedding model, UMAP configuration,
clustering resolution) require recalibration for each new stakeholder
group and discourse domain.

\subsection{Conclusion}\label{conclusion}

This study provides a measurement framework for auditing stakeholder
representation in technology discourse. We move beyond the normative
claim that ``artists should be heard'' to a measurable demonstration
that, in the current public sphere, artist concerns are compressed into
a narrow region of meaning-space while institutional, technical, and
spectacular narratives occupy the broader discursive landscape.

The compression is structural, not incidental. It operates through
topical exclusion, frame redirection into aesthetic discourse, binary
simplification of labour claims, and voice collapse across substantively
distinct positions. At a macro level, these patterns combine in the
operation of media framing, algorithmic amplification, and event-driven
spectacle (Pane, 2025; Cheung, 2025). For AI governance, relying on
public discourse as a proxy for stakeholder priorities risks legislating
against a compressed signal rather than the underlying distribution of
concerns. The dimensions whose internal political complexity matters
most for policy (ownership, transparency, compensation), alongside
experimental dimensions like utility, are flattened most severely.

The public sphere has captured the issue of AI and art. It has not
captured the reality of artists' concerns.

\subsection{References}\label{references}

Akrich M (1992) The de-scription of technical objects. In: Bijker WE and
Law J (eds) Shaping Technology/Building Society: Studies in
Sociotechnical Change. Cambridge, MA: MIT Press, pp.~205--224.

Andersen et al.~v. Stability AI et al.~(2023) Class action complaint.
U.S. District Court, Northern District of California, Case
No.~3:23-cv-00201.

Banks J and Li J (2025) Wherefore art thou: Mapping public debates about
image-generative AI. In: Proceedings of the 58th Hawaii International
Conference on System Sciences. DOI: 10.24251/HICSS.2025.067.

Birhane A (2021) Algorithmic injustice: A relational ethics approach.
Patterns 2(2): 100205. DOI: 10.1016/j.patter.2021.100205.

Bishop S (2025) Influencer creep: How artists strategically navigate the
platformisation of art worlds. New Media \& Society 27(4): 2109--2126.
DOI: 10.1177/14614448231206090.

Bøgh LS (2025) Framing the machine: A comparative analysis of AI-art
discourse in Danish news media. Master's thesis, Malmö University.
Available at:
https://mau.diva-portal.org/smash/get/diva2:1981056/FULLTEXT01.pdf
(accessed 17 March 2026).

Bowker GC and Star SL (1999) Sorting Things Out: Classification and Its
Consequences. Cambridge, MA: MIT Press.

Brause SR, Schäfer MS, Katzenbach C, Mao Y, Richter V and Zeng J (2025)
Sociotechnical imaginaries and public communication: Analytical
framework and empirical illustration using the case of artificial
intelligence. Convergence: The International Journal of Research into
New Media Technologies 31(4): 1267--1287. DOI:
10.1177/13548565251338192.

Browne K (2022) Who (or what) is an AI artist? Leonardo 55(2): 130--134.
DOI: 10.1162/leon\_a\_02092.

Campos Valverde R and Kaye DBV (2026) `A safe, responsible, and
profitable ecosystem of music': Analyzing perceptions and implementation
of generative AI in the music industry. New Media \& Society. Epub ahead
of print. DOI: 10.1177/14614448251411526.

Cheung S (2025) Generative AI, generating crisis: Framing opportunity
and threat in AI governance. Information, Communication \& Society. Epub
ahead of print. DOI: 10.1080/1369118X.2025.2542356.

Cohen M, Khavkin M, Movsowitz Davidow D, et al.~(2026) ChatGPT in the
public eye: Ethical principles and generative concerns in social media
discussions. New Media \& Society 28(1): 5--31. DOI:
10.1177/14614448241279034.

Crawford K (2021) Atlas of AI: Power, Politics, and the Planetary Costs
of Artificial Intelligence. New Haven: Yale University Press.

Creative Rights in AI Coalition (2024) Statement on creative rights in
AI. Launched 16 December. Available at:
https://www.creativerightsinai.co.uk/ (accessed 17 March 2026).

Creators Coalition on AI (2025) US creators coalition on AI. Los Angeles
Times, 17 December. Available at:
https://www.latimes.com/entertainment-arts/business/story/2025-12-17/hollywood-stars-launch-creators-coalition-on-ai
(accessed 17 March 2026).

D'Ignazio C and Klein LF (2020) Data Feminism. Cambridge, MA: MIT Press.

DiMaggio P, Nag M and Blei D (2013) Exploiting affinities between topic
modeling and the sociological perspective on culture: Application to
newspaper coverage of U.S. government arts funding. Poetics 41(6):
570--606. DOI: 10.1016/j.poetic.2013.08.004.

Elmholdt KT, Nielsen JA, Florczak CK, et al.~(2025) The hopes and fears
of artificial intelligence: A comparative computational discourse
analysis. AI \& Society 40(6): 4765--4782. DOI:
10.1007/s00146-025-02214-z.

Entman RM (1993) Framing: Toward clarification of a fractured paradigm.
Journal of Communication 43(4): 51--58. DOI:
10.1111/j.1460-2466.1993.tb01304.x.

Ferrari F, van Dijck J and van den Bosch A (2025) Observe, inspect,
modify: Three conditions for generative AI governance. New Media \&
Society 27(5): 2788--2806. DOI: 10.1177/14614448231214811.

Fricker M (2007) Epistemic Injustice: Power and the Ethics of Knowing.
Oxford: Oxford University Press. DOI:
10.1093/acprof:oso/9780198237907.001.0001.

Goetze TS (2024) AI art is theft: Labour, extraction, and exploitation,
or, on the dangers of stochastic Pollocks. In: Proceedings of the 2024
ACM Conference on Fairness, Accountability, and Transparency (FAccT
'24). New York: ACM. DOI: 10.1145/3630106.3658898.

Goodfellow I, Pouget-Abadie J, Mirza M, et al.~(2014) Generative
adversarial nets. In: Advances in Neural Information Processing Systems,
vol.~27, pp.~2672--2680.

Grootendorst M (2022) BERTopic: Neural topic modeling with a class-based
TF-IDF procedure. arXiv: 2203.05794.

Huang X and Gadavanij S (2025) Power and marginalization in discourse on
AI in education (AIEd): Social actors' representation in China Daily
(2018--2023). Humanities and Social Sciences Communications 12: article
412. DOI: 10.1057/s41599-025-04621-5.

Jasanoff S (2015) Future imperfect: Science, technology, and the
imaginations of modernity. In: Jasanoff S and Kim S-H (eds) Dreamscapes
of Modernity: Sociotechnical Imaginaries and the Fabrication of Power.
Chicago: University of Chicago Press, pp.~1-33.

Jiang HH, Brown L, Cheng J, et al.~(2023) AI art and its impact on
artists. In: Proceedings of the AAAI/ACM Conference on AI, Ethics, and
Society. New York: ACM, pp.~363--374. DOI: 10.1145/3600211.3604681.

Joyce K and Cruz TM (2024) A sociology of artificial intelligence:
Inequalities, power, and data justice. Socius: Sociological Research for
a Dynamic World 10: 1--12. DOI: 10.1177/23780231241275393.

Kawakami R and Venkatagiri S (2024) The impact of generative AI on
artists. In: Proceedings of the 16th Conference on Creativity \&
Cognition. New York: ACM, pp.~79--82. DOI: 10.1145/3635636.3664263.

Kay J, Kasirzadeh A and Mohamed S (2024) Epistemic injustice in
generative AI. In: Proceedings of the 2024 AAAI/ACM Conference on AI,
Ethics, and Society. New York: ACM. arXiv: 2408.11441.

Li L, Vásárhelyi O and Vedres B (2024) Social bots spoil activist
sentiment without eroding engagement. Scientific Reports 14: 27005. DOI:
10.1038/s41598-024-74032-0.

Lovato J, Zimmerman JW, Smith I, et al.~(2024) Foregrounding artist
opinions: A survey study on transparency, ownership, and fairness in AI
generative art. In: Proceedings of the AAAI/ACM Conference on AI,
Ethics, and Society, vol.~7(1). New York: ACM, pp.~905--916. DOI:
10.1609/aies.v7i1.31691.

MacKenzie D (2006) An Engine, Not a Camera: How Financial Models Shape
Markets. Cambridge, MA: MIT Press.

Małecki WP, Messingschlager TV and Appel M (2025) The impact of exposure
to generative AI art on aesthetic appreciation, perceptions of AI mind,
and evaluations of AI and of art careers. New Media \& Society 27(9):
5410--5432. DOI: 10.1177/14614448251344590.

Matsui A and Ferrara E (2024) Word embedding for social sciences: An
interdisciplinary survey. PeerJ Computer Science 10: e2562. DOI:
10.7717/peerj-cs.2562.

Mazzone M and Elgammal A (2019) Art, creativity, and the potential of
artificial intelligence. Arts 8(1): 26. DOI: 10.3390/arts8010026.

McCormack J, Gifford T and Hutchings P (2019) Autonomy, authenticity,
authorship and intention in computer generated art. In: Ekárt A, Liapis
A and Castro Pena ML (eds) Computational Intelligence in Music, Sound,
Art, and Design (EvoMUSART 2019). Lecture Notes in Computer Science,
vol.~11453. Cham: Springer, pp.~35--50. DOI:
10.1007/978-3-030-16667-0\_3.

McHugh ML (2012) Interrater reliability: the kappa statistic. Biochemia
Medica 22(3): 276--282. DOI: 10.11613/BM.2012.031.

Mordvintsev A, Olah C and Tyka M (2015) Inceptionism: Going deeper into
neural networks. Google Research Blog, 18 June. Available at:
https://research.google/blog/inceptionism-going-deeper-into-neural-networks/
(accessed 17 March 2026).

Noble SU (2018) Algorithms of Oppression: How Search Engines Reinforce
Racism. New York: NYU Press.

Pane S (2025) La esfera pública `post-digital' europea: fundamentos de
un paradigma emergente en las ciencias sociales {[}The European
`post-digital' public sphere: Foundations of an emerging paradigm in the
social sciences{]}. methaodos. revista de ciencias sociales 13(1). DOI:
10.17502/mrcs.v13i1.866.

Ramesh A, Pavlov M, Goh G, et al.~(2021) Zero-shot text-to-image
generation. In: Proceedings of the 38th International Conference on
Machine Learning (PMLR), vol.~139, pp.~8821--8831. Available at:
https://proceedings.mlr.press/v139/ramesh21a.html (accessed 17 March
2026).

Rombach R, Blattmann A, Lorenz D, et al.~(2022) High-resolution image
synthesis with latent diffusion models. In: Proceedings of the IEEE/CVF
Conference on Computer Vision and Pattern Recognition, pp.~10684--10695.
DOI: 10.1109/CVPR52688.2022.01042.

Saeedi M and Taleghani M (2025) Uncovering the discourses around the
diffusion of generative art: A topic modeling approach. In: Proceedings
of the Americas Conference on Information Systems.

Saurwein F, Brantner C and Möck L (2025) Responsibility networks in
media discourses on automation: A comparative analysis of social media
algorithms and social companions. New Media \& Society 27(3):
1752--1773. DOI: 10.1177/14614448231203310.

Sichach M (2024) From mainstream media to algorithms: Agenda setting in
the age of artificial intelligence. SSRN Working Paper, 5 October.
Available at: https://ssrn.com/abstract=5040125 (accessed 17 March
2026).

Suchman L (2002) Located accountabilities in technology production.
Scandinavian Journal of Information Systems 14(2): 91--105.

Suresh H, Tseng E, Young M, Gray ML, Pierson E and Levy K (2024)
Participation in the age of foundation models. In: Proceedings of the
2024 ACM Conference on Fairness, Accountability, and Transparency (FAccT
'24). New York: ACM, pp.~1609--1621. DOI: 10.1145/3630106.3658992.

Varvasovszky Z and Brugha R (2000) A stakeholder analysis. Health Policy
and Planning 15(3): 338--345. DOI: 10.1093/heapol/15.3.338.

Wald R, Araujo T, van Oosten JMF and Piotrowski JT (2026) What does the
news say about Siri, Alexa, and co.? A topic model network analysis of
Dutch news messages about virtual assistants between 2011 and 2022. New
Media \& Society. Epub ahead of print. DOI: 10.1177/14614448251413694.

Widder DG (2024) Epistemic power in AI ethics labor: Legitimizing
located complaints. In: Proceedings of the 2024 ACM Conference on
Fairness, Accountability, and Transparency (FAccT '24). New York: ACM.
DOI: 10.1145/3630106.3658973.

Zai F, Rohrbach T and Hänggli Fricker R (2025) Voices and media frames
in the public debate on artificial intelligence: Comparing results from
manual and automated content analysis. Frontiers in Communication 10:
article 1599854. DOI: 10.3389/fcomm.2025.1599854.

Ziems C, Held W, Shaikh O, et al.~(2024) Can large language models
transform computational social science? Computational Linguistics 50(1):
237--291. DOI: 10.1162/coli\_a\_00502.

\subsection{Data availability}\label{data-availability}

The public discourse corpus generated during the current study, together
with all derived artifacts (cleaned chunks, public probes, probe
embeddings, and consensus coordinates), is available in the project
GitHub repository
(https://github.com/AArtist1/When-Algorithms-Meet-Artists). The artist
survey data analysed during the current study are available through
Lovato et al.~(2024).

\subsection{Code availability}\label{code-availability}

All analysis code, including the consensus UMAP pipeline, probe
extraction scripts, and statistical tests, is available at
https://github.com/AArtist1/When-Algorithms-Meet-Artists under GNU GPL
v3.

\subsection{Author contributions}\label{author-contributions}

A.M.-G. and O.M. contributed equally to this work. A.M.-G. conceived the
research question, curated the public discourse corpus, organised artist
survey data into thematic dimensions, provided domain expertise as a
practising artist, and contributed to interpretation. O.M. designed and
implemented the computational pipeline, conducted statistical analyses,
and contributed to interpretation. Both authors contributed equally to
writing the manuscript.

\subsection{Competing interests}\label{competing-interests}

The authors declare no competing financial or non-financial interests in
relation to the work described.

\subsection{Ethics statement}\label{ethics-statement}

This study analyses publicly available documents and previously
published survey data (Lovato et al., 2024). No new human subjects data
were collected. The Lovato et al.~survey data were collected under their
institution's ethical approval.

\subsection{Acknowledgements}\label{acknowledgements}

We thank Lovato et al.~for making their survey data available. AI-human
collaboration during manuscript preparation was documented using the
TRACE protocol. Machine-readable provenance logs are available in the
project repository under \texttt{trace\_logs/}.

\newpage

\section{Supplementary Information}\label{supplementary-information}

\textbf{When Algorithms Meet Artists: Semantic Compression and
Stakeholder Marginalisation in Public AI-Art Discourse}

\subsection{Table S1: Search Queries}\label{table-s1-search-queries}

Public discourse documents were identified via Google Search using 17
targeted queries. For each query, all unique documents appearing on the
first page of results were sampled. Searches were conducted in a
logged-out browser environment to reduce personalisation artifacts.

\begin{longtable}[]{@{}
  >{\raggedright\arraybackslash}p{(\linewidth - 2\tabcolsep) * \real{0.5889}}
  >{\raggedright\arraybackslash}p{(\linewidth - 2\tabcolsep) * \real{0.3889}}@{}}
\toprule\noalign{}
\begin{minipage}[b]{\linewidth}\raggedright
\textbf{Search Phrase}
\end{minipage} & \begin{minipage}[b]{\linewidth}\raggedright
\textbf{Relevant Documents in Top 10}
\end{minipage} \\
\midrule\noalign{}
\endhead
\bottomrule\noalign{}
\endlastfoot
AI art & 4 \\
AI generated art & 5 \\
AI art impact on artists & 7 \\
Effects of AI image generators on artists' careers & 5 \\
Artists' responses to generative AI & 7 \\
Artists' concerns about AI art & 5 \\
Interviews with artists about AI image generators & 8 \\
How AI affects freelance artists & 9 \\
AI copyright challenges for artists & 12 \\
Artists suing AI companies & 10 \\
How artists adapt to AI tools & 6 \\
Artists protest AI training data usage & 7 \\
Artists' opinion on DeepDream art & 10 \\
AI art exhibition & 8 \\
Art incorporating AI & 11 \\
Creatives and AI & 9 \\
AI and the creative job market & 11 \\
\textbf{\emph{Total}} & \textbf{\emph{134}} \\
\end{longtable}

After deduplication across queries and removal of 3 duplicate articles
(same content ingested as both text and audio transcript), the final
corpus comprised 125 unique documents spanning 2013-2025.

\subsection{Table S2: Full Topic
Inventory}\label{table-s2-full-topic-inventory}

20 topics identified via KMeans clustering (k=20, selected by 4-stage
hyperparameter validation with topic quality prioritisation) of the
consensus UMAP embedding. Labels assigned by multi-annotator panel (2
human coders + 4 LLMs: Claude Opus 4.6, Claude Sonnet 4.6, GPT-5.4-mini,
GPT-5.4-nano) converge on as the most defensible reading of each
cluster's chunk contents. These are interpretive outputs, not
ground-truth descriptions of the topics. The H3 compression findings
depend on the gap between these interpretations and where artist probes
actually land.

\subsubsection{Macro-thematic groupings}\label{macro-thematic-groupings}

\begin{longtable}[]{@{}
  >{\raggedright\arraybackslash}p{(\linewidth - 4\tabcolsep) * \real{0.3253}}
  >{\raggedright\arraybackslash}p{(\linewidth - 4\tabcolsep) * \real{0.2651}}
  >{\raggedright\arraybackslash}p{(\linewidth - 4\tabcolsep) * \real{0.3855}}@{}}
\toprule\noalign{}
\begin{minipage}[b]{\linewidth}\raggedright
\textbf{Macro-theme}
\end{minipage} & \begin{minipage}[b]{\linewidth}\raggedright
\textbf{Share of Corpus}
\end{minipage} & \begin{minipage}[b]{\linewidth}\raggedright
\textbf{Topics}
\end{minipage} \\
\midrule\noalign{}
\endhead
\bottomrule\noalign{}
\endlastfoot
Philosophy of Creativity & 38.1\% & 1, 4, 11, 13, 17 \\
Practice and Pedagogy & 34.8\% & 3, 7, 14, 16 \\
Technical Genealogy & 14.7\% & 2, 5, 15, 19 \\
Governance and Rights & 7.5\% & 8, 10, 18 \\
Institutions and Markets & 4.9\% & 0, 6, 9, 12 \\
\end{longtable}

\subsubsection{\texorpdfstring{\protect\includegraphics[width=6.5in,height=5.11111in]{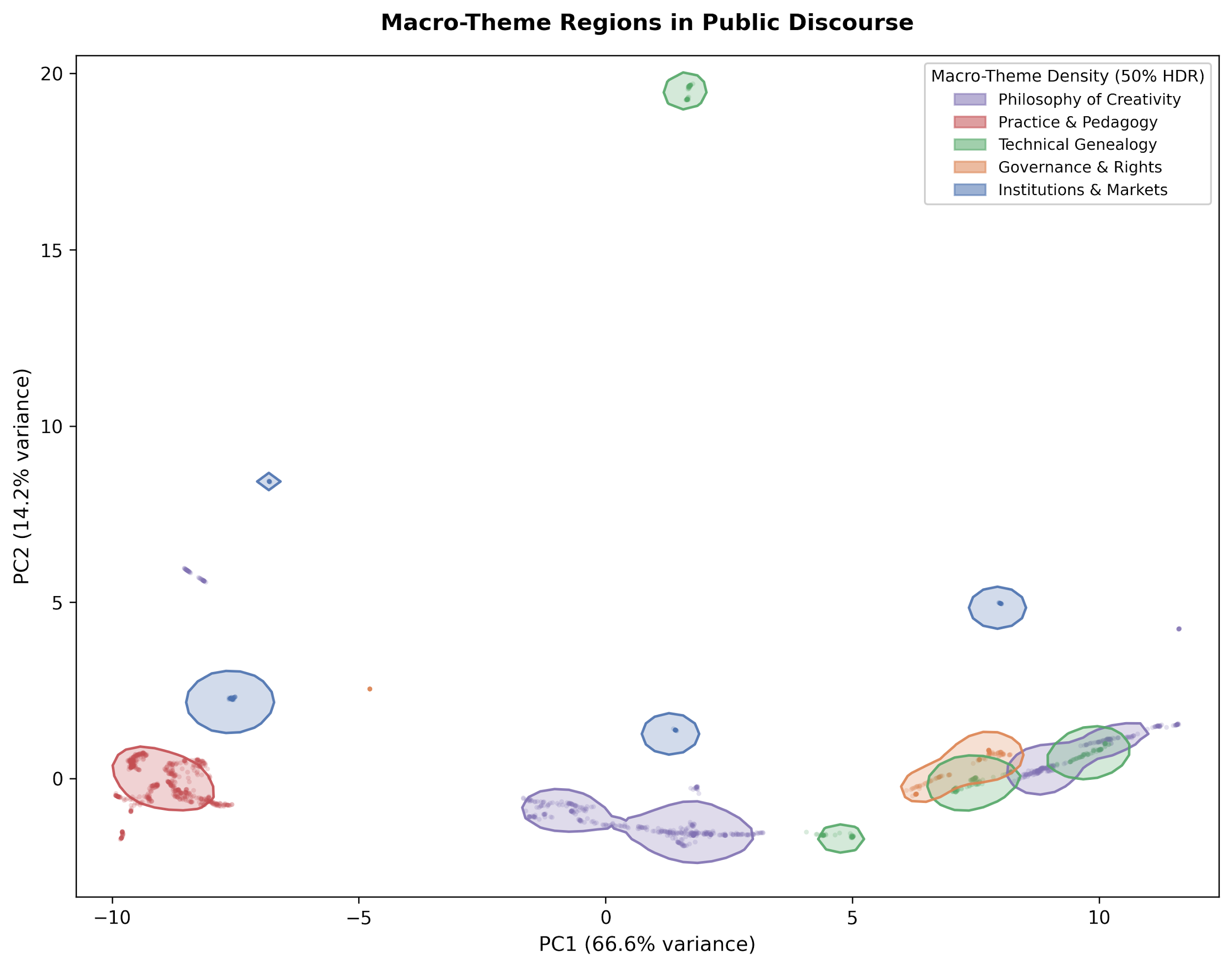}}{}}\label{section}

\emph{\textbf{Figure S1.} Semantic map of public discourse macro-theme
regions (50\% highest-density contours on the 2D PCA projection of the
5-dimensional consensus UMAP coordinates). Each coloured region encloses
the densest 50\% of chunks belonging to one of the five macro-thematic
groups identified via KMeans clustering and multi-annotator
interpretation. The spatial dispersion of public discourse contrasts
with the tight concentration of artist probes shown in Figure 1.}

\subsubsection{Full topic inventory}\label{full-topic-inventory}

\begin{longtable}[]{@{}
  >{\raggedright\arraybackslash}p{(\linewidth - 6\tabcolsep) * \real{0.0896}}
  >{\raggedright\arraybackslash}p{(\linewidth - 6\tabcolsep) * \real{0.3731}}
  >{\raggedright\arraybackslash}p{(\linewidth - 6\tabcolsep) * \real{0.3209}}
  >{\raggedright\arraybackslash}p{(\linewidth - 6\tabcolsep) * \real{0.2015}}@{}}
\toprule\noalign{}
\begin{minipage}[b]{\linewidth}\raggedright
\textbf{Topic}
\end{minipage} & \begin{minipage}[b]{\linewidth}\raggedright
\textbf{Panel interpretation}
\end{minipage} & \begin{minipage}[b]{\linewidth}\raggedright
\textbf{Top Keywords}
\end{minipage} & \begin{minipage}[b]{\linewidth}\raggedright
\textbf{Macro-theme}
\end{minipage} \\
\midrule\noalign{}
\endhead
\bottomrule\noalign{}
\endlastfoot
0 & Decentralized Infrastructure for Art Ecosystems & decentralized,
technologies, cultural & Institutions and Markets \\
1 & AI as Creative Collaborator & ai, art, human, data & Philosophy of
Creativity \\
2 & Machine Learning Art Theory and Practice & MIT, machine learning,
Microsoft & Technical Genealogy \\
3 & Personal Reflections on AI and Art & know, ai, art, really, yeah &
Practice and Pedagogy \\
4 & Mental Models and Abstract Thought & sort, kind, mental images,
language & Philosophy of Creativity \\
5 & Harold Cohen and AARON Legacy & Cohen, AARON, computer, creativity &
Technical Genealogy \\
6 & Future of Arts Journalism and Museums & museums, publishing,
writing, journalism & Institutions and Markets \\
7 & Conversational Reflections on Art Practice & know, art, really,
kind, lot & Practice and Pedagogy \\
8 & AI Copyright and Legal Protection & copyright, coders, software, law
& Governance and Rights \\
9 & Digital Art Exhibition and Display & internet, MoMA, exhibition &
Institutions and Markets \\
10 & Artist Defense Tools Against AI & Nightshade, Glaze, tools &
Governance and Rights \\
11 & AI Authorship and Creative Agency & Chung, authorship, agency &
Philosophy of Creativity \\
12 & Media Coverage and AI Panic & ai, effective altruism, media,
humanity & Institutions and Markets \\
13 & AI Art Authenticity and Human Creativity & ai, art, artists, image,
human & Philosophy of Creativity \\
14 & Informal AI Creative Tool Discourse & gonna, students, Adobe,
Firefly & Practice and Pedagogy \\
15 & Deep Dream and Neural Network Visualization & Mordvintsev, neural,
deep, Deep Dream & Technical Genealogy \\
16 & Artist Reflections on Technology & know, art, really, artists &
Practice and Pedagogy \\
17 & Artist-Centered AI Design and Ethics & prompts, design, ethics &
Philosophy of Creativity \\
18 & AI Art Authorship and Copyright Debates & Midjourney, Allen,
copyright & Governance and Rights \\
19 & Generative Art History and Pioneers & ai art, generative art,
Barrat & Technical Genealogy \\
\end{longtable}

\subsection{Table S3: Likert Anchor Design
Matrix}\label{table-s3-likert-anchor-design-matrix}

250 synthetic Likert-style anchor statements were generated using
GPT-5.4-mini following a 5 x 5 x 10 factorial design: 5 themes x 5
agreement levels x 10 discourse styles.

These anchors were used as retrieval queries to extract style-matched
public probes from the public corpus. The anchors themselves were
discarded after extraction and are not used in any downstream analysis.

\textbf{Themes (5):} utility, ownership, transparency, threat,
compensation

\textbf{Agreement Levels (5):} strongly disagree, disagree, neutral,
agree, strongly agree

\textbf{Discourse Styles (10):} blog opinion, news editorial, artist
interview, academic article, policy report, panel discussion, first
person, union statement, gallery statement, research summary

\subsubsection{Representative examples (one per theme x
level)}\label{representative-examples-one-per-theme-x-level}

\begin{longtable}[]{@{}
  >{\raggedright\arraybackslash}p{(\linewidth - 6\tabcolsep) * \real{0.0743}}
  >{\raggedright\arraybackslash}p{(\linewidth - 6\tabcolsep) * \real{0.0990}}
  >{\raggedright\arraybackslash}p{(\linewidth - 6\tabcolsep) * \real{0.7079}}
  >{\raggedright\arraybackslash}p{(\linewidth - 6\tabcolsep) * \real{0.1089}}@{}}
\toprule\noalign{}
\begin{minipage}[b]{\linewidth}\raggedright
\textbf{Theme}
\end{minipage} & \begin{minipage}[b]{\linewidth}\raggedright
\textbf{Level}
\end{minipage} & \begin{minipage}[b]{\linewidth}\raggedright
\textbf{Example}
\end{minipage} & \begin{minipage}[b]{\linewidth}\raggedright
\textbf{Style}
\end{minipage} \\
\midrule\noalign{}
\endhead
\bottomrule\noalign{}
\endlastfoot
Utility & Strongly disagree & ``From my perspective, artificial
intelligence art models are ruining the field of art rather than
improving it.'' & blog opinion \\
Utility & Disagree & ``The panel expressed skepticism that artificial
intelligence art models truly benefit the field of art.'' & panel
discussion \\
Utility & Neutral & ``The report takes no position on whether artificial
intelligence art models are good or bad for art.'' & research report \\
Utility & Agree & ``The article argues that artificial intelligence art
models open exciting new possibilities for artists.'' & news opinion \\
Utility & Strongly agree & ``This manifesto celebrates artificial
intelligence art models as a transformative good for creative
practice.'' & manifesto \\
Ownership & Strongly disagree & ``In my view, artificial intelligence
generated artwork in an artist's style should not be considered the
artist's property at all.'' & legal opinion \\
Ownership & Neutral & ``The article maps the legal uncertainties around
ownership of artificial intelligence generated images in an artist's
style.'' & academic article \\
Ownership & Strongly agree & ``The manifesto declares that artificial
intelligence generated images in an artist's style belong entirely to
that artist.'' & manifesto \\
Transparency & Strongly disagree & ``Our company rejects proposals to
mandate detailed transparency about the art and images used to train our
artificial intelligence systems.'' & corporate statement \\
Transparency & Strongly agree & ``The report strongly recommends strict
disclosure mandates covering all art and images used to train artificial
intelligence models.'' & policy report \\
Threat & Strongly disagree & ``The paper contends that artificial
intelligence art models complement, rather than threaten, human artistic
labour.'' & academic article \\
Threat & Strongly agree & ``Our statement declares that artificial
intelligence art models are a direct attack on working artists and their
jobs.'' & union statement \\
Compensation & Strongly disagree & ``The report concludes that mandatory
artist compensation for artificial intelligence training would be
inappropriate.'' & policy report \\
Compensation & Strongly agree & ``This manifesto demands strong profit
sharing and royalty rights for artists in all artificial intelligence
art contexts.'' & manifesto \\
\end{longtable}

The full 250-statement set is available as likert\_anchor\_phrases.csv
in the project repository.

\subsection{Table S4: Artist Frame
Distributions}\label{table-s4-artist-frame-distributions}

Data from 252 US-based practising artists filtered from the Lovato et
al.~(2024) survey (n = 459 total). Response categories follow Lovato et
al.'s coding: Agree (Strongly Agree + Agree), Neutral, Disagree
(Strongly Disagree + Disagree).

\includegraphics[width=6.5in,height=5.11111in]{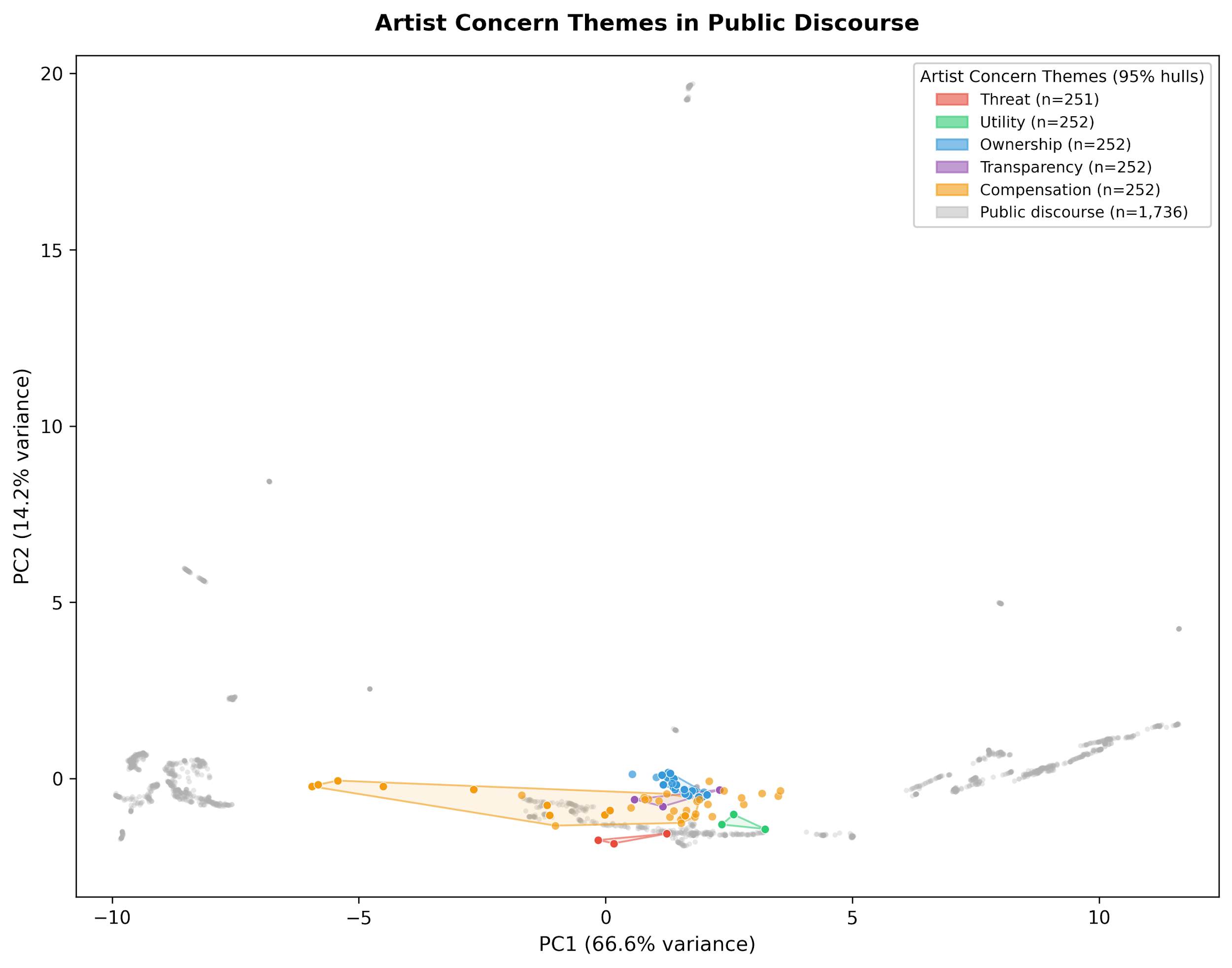}

\emph{\textbf{Figure S2.} Artist concern themes projected into public
discourse space (95\% convex hulls on the 2D PCA projection of the
5-dimensional consensus UMAP coordinates). Each coloured polygon
encloses 95\% of the probes for one concern dimension (threat, utility,
ownership, transparency, compensation). The compensation theme shows the
greatest spatial spread, while the other four themes collapse into a
tight central region, visually confirming the voice-collapse mechanism:
substantively distinct concerns occupy nearly the same coordinates when
artist voices share a first-person practice register. Grey points in the
background represent the 1,736 public discourse chunks, illustrating how
narrowly the artist themes occupy the broader discourse space.}

\subsubsection{Dimension distributions}\label{dimension-distributions}

\textbf{Threat} (``I see AI models as a threat to art workers''):

\begin{longtable}[]{@{}
  >{\raggedright\arraybackslash}p{(\linewidth - 4\tabcolsep) * \real{0.2667}}
  >{\raggedright\arraybackslash}p{(\linewidth - 4\tabcolsep) * \real{0.3333}}
  >{\raggedright\arraybackslash}p{(\linewidth - 4\tabcolsep) * \real{0.3733}}@{}}
\toprule\noalign{}
\begin{minipage}[b]{\linewidth}\raggedright
\textbf{Response}
\end{minipage} & \begin{minipage}[b]{\linewidth}\raggedright
\textbf{n}
\end{minipage} & \begin{minipage}[b]{\linewidth}\raggedright
\textbf{\%}
\end{minipage} \\
\midrule\noalign{}
\endhead
\bottomrule\noalign{}
\endlastfoot
Agree & 152 & 60.6\% \\
Disagree & 55 & 21.9\% \\
Neutral & 44 & 17.5\% \\
\end{longtable}

\textbf{Utility} (``I see AI art models as a positive development in the
field of art''):

\begin{longtable}[]{@{}
  >{\raggedright\arraybackslash}p{(\linewidth - 4\tabcolsep) * \real{0.2703}}
  >{\raggedright\arraybackslash}p{(\linewidth - 4\tabcolsep) * \real{0.3243}}
  >{\raggedright\arraybackslash}p{(\linewidth - 4\tabcolsep) * \real{0.3784}}@{}}
\toprule\noalign{}
\begin{minipage}[b]{\linewidth}\raggedright
\textbf{Response}
\end{minipage} & \begin{minipage}[b]{\linewidth}\raggedright
\textbf{n}
\end{minipage} & \begin{minipage}[b]{\linewidth}\raggedright
\textbf{\%}
\end{minipage} \\
\midrule\noalign{}
\endhead
\bottomrule\noalign{}
\endlastfoot
Agree & 116 & 46.0\% \\
Disagree & 69 & 27.4\% \\
Neutral & 67 & 26.6\% \\
\end{longtable}

\textbf{Transparency} (``Model creators should be required to disclose
in detail what art and images they use to train their AI models''):

\begin{longtable}[]{@{}
  >{\raggedright\arraybackslash}p{(\linewidth - 4\tabcolsep) * \real{0.2667}}
  >{\raggedright\arraybackslash}p{(\linewidth - 4\tabcolsep) * \real{0.3333}}
  >{\raggedright\arraybackslash}p{(\linewidth - 4\tabcolsep) * \real{0.3733}}@{}}
\toprule\noalign{}
\begin{minipage}[b]{\linewidth}\raggedright
\textbf{Response}
\end{minipage} & \begin{minipage}[b]{\linewidth}\raggedright
\textbf{n}
\end{minipage} & \begin{minipage}[b]{\linewidth}\raggedright
\textbf{\%}
\end{minipage} \\
\midrule\noalign{}
\endhead
\bottomrule\noalign{}
\endlastfoot
Agree & 204 & 81.0\% \\
Neutral & 25 & 9.9\% \\
Disagree & 23 & 9.1\% \\
\end{longtable}

\textbf{Ownership} (``Should AI-generated artwork in your style be
considered your property as the original artist?''):

\begin{longtable}[]{@{}
  >{\raggedright\arraybackslash}p{(\linewidth - 4\tabcolsep) * \real{0.3929}}
  >{\raggedright\arraybackslash}p{(\linewidth - 4\tabcolsep) * \real{0.2500}}
  >{\raggedright\arraybackslash}p{(\linewidth - 4\tabcolsep) * \real{0.3333}}@{}}
\toprule\noalign{}
\begin{minipage}[b]{\linewidth}\raggedright
\textbf{Response}
\end{minipage} & \begin{minipage}[b]{\linewidth}\raggedright
\textbf{n}
\end{minipage} & \begin{minipage}[b]{\linewidth}\raggedright
\textbf{\%}
\end{minipage} \\
\midrule\noalign{}
\endhead
\bottomrule\noalign{}
\endlastfoot
Agree (artist owns) & 110 & 43.7\% \\
Disagree (artist does not own) & 91 & 36.1\% \\
Neutral & 51 & 20.2\% \\
\end{longtable}

\textbf{Compensation} (``What type of compensation would you consider
fair?''):

\begin{longtable}[]{@{}
  >{\raggedright\arraybackslash}p{(\linewidth - 4\tabcolsep) * \real{0.4457}}
  >{\raggedright\arraybackslash}p{(\linewidth - 4\tabcolsep) * \real{0.2283}}
  >{\raggedright\arraybackslash}p{(\linewidth - 4\tabcolsep) * \real{0.3043}}@{}}
\toprule\noalign{}
\begin{minipage}[b]{\linewidth}\raggedright
\textbf{Stance}
\end{minipage} & \begin{minipage}[b]{\linewidth}\raggedright
\textbf{n}
\end{minipage} & \begin{minipage}[b]{\linewidth}\raggedright
\textbf{\%}
\end{minipage} \\
\midrule\noalign{}
\endhead
\bottomrule\noalign{}
\endlastfoot
No compensation, but no for-profit use & 56 & 22.2\% \\
Would donate freely & 50 & 19.8\% \\
Revenue share from model creators & 27 & 10.7\% \\
Rejects all listed options & 25 & 9.9\% \\
Flat fee & 22 & 8.7\% \\
Revenue share from any profit & 20 & 7.9\% \\
Revenue share from derivatives & 17 & 6.7\% \\
No profit needed & 13 & 5.2\% \\
No profit for anyone & 12 & 4.8\% \\
Other & 8 & 3.2\% \\
AI tax & 2 & 0.8\% \\
\end{longtable}

\subsubsection{Threat x utility
cross-tabulation}\label{threat-x-utility-cross-tabulation}

This cross-tab reveals that the media binary (``AI is good'' vs ``AI is
bad'' for art) can accommodate only 41.3\% of artist positions. 58.7\%
of artists hold positions that require both categories simultaneously or
neither.

\begin{longtable}[]{@{}
  >{\raggedright\arraybackslash}p{(\linewidth - 8\tabcolsep) * \real{0.2170}}
  >{\raggedright\arraybackslash}p{(\linewidth - 8\tabcolsep) * \real{0.1981}}
  >{\raggedright\arraybackslash}p{(\linewidth - 8\tabcolsep) * \real{0.2264}}
  >{\raggedright\arraybackslash}p{(\linewidth - 8\tabcolsep) * \real{0.2170}}
  >{\raggedright\arraybackslash}p{(\linewidth - 8\tabcolsep) * \real{0.1226}}@{}}
\toprule\noalign{}
\begin{minipage}[b]{\linewidth}\raggedright
\end{minipage} & \begin{minipage}[b]{\linewidth}\raggedright
\textbf{Utility: Agree}
\end{minipage} & \begin{minipage}[b]{\linewidth}\raggedright
\textbf{Utility: Disagree}
\end{minipage} & \begin{minipage}[b]{\linewidth}\raggedright
\textbf{Utility: Neutral}
\end{minipage} & \begin{minipage}[b]{\linewidth}\raggedright
\textbf{Total}
\end{minipage} \\
\midrule\noalign{}
\endhead
\bottomrule\noalign{}
\endlastfoot
\textbf{Threat: Agree} & \textbf{\emph{55 (21.8\%)}} & 60 (23.8\%) & 37
(14.7\%) & 152 \\
\textbf{Threat: Disagree} & 44 (17.5\%) & 3 (1.2\%) & 8 (3.2\%) & 55 \\
\textbf{Threat: Neutral} & 16 (6.3\%) & 6 (2.4\%) & 22 (8.7\%) & 44 \\
\end{longtable}

Cells in bold represent the ``dual'' position (agree that AI is both a
threat AND a positive development): 55 artists (21.8\%), consistent with
Lovato et al.'s (2024) finding that 22\% of their full sample held
``complex opinions'' spanning both dimensions.

Artists who fit the pure media binary: - ``AI is bad'' (Threat: Agree +
Utility: Disagree): 60 artists (23.8\%) - ``AI is good'' (Threat:
Disagree + Utility: Agree): 44 artists (17.5\%) - \textbf{Total fitting
binary: 104 (41.3\%)} - \textbf{Total NOT fitting binary: 148 (58.7\%)}

\subsubsection{AI experience and position
complexity}\label{ai-experience-and-position-complexity}

Artists who have used AI art models are nearly three times more likely
to hold the dual threat-and-utility position than those who have not:

\begin{longtable}[]{@{}
  >{\raggedright\arraybackslash}p{(\linewidth - 8\tabcolsep) * \real{0.3030}}
  >{\raggedright\arraybackslash}p{(\linewidth - 8\tabcolsep) * \real{0.2020}}
  >{\raggedright\arraybackslash}p{(\linewidth - 8\tabcolsep) * \real{0.1515}}
  >{\raggedright\arraybackslash}p{(\linewidth - 8\tabcolsep) * \real{0.1212}}
  >{\raggedright\arraybackslash}p{(\linewidth - 8\tabcolsep) * \real{0.2020}}@{}}
\toprule\noalign{}
\begin{minipage}[b]{\linewidth}\raggedright
\end{minipage} & \begin{minipage}[b]{\linewidth}\raggedright
\textbf{Dual position}
\end{minipage} & \begin{minipage}[b]{\linewidth}\raggedright
\textbf{Not dual}
\end{minipage} & \begin{minipage}[b]{\linewidth}\raggedright
\textbf{Total}
\end{minipage} & \begin{minipage}[b]{\linewidth}\raggedright
\textbf{\% Dual}
\end{minipage} \\
\midrule\noalign{}
\endhead
\bottomrule\noalign{}
\endlastfoot
\textbf{Have used AI models} & 44 & 104 & 148 & 29.7\% \\
\textbf{Have not used AI models} & 11 & 90 & 101 & 10.9\% \\
\end{longtable}

Chi-square = 11.31, p \textless{} 0.001, Cramer's V = 0.21. Fisher's
exact test confirms (odds ratio = 3.46, p \textless{} 0.001).

\subsubsection{Cross-dimensional
combinations}\label{cross-dimensional-combinations}

Across all five dimensions, the 252 artists produced 137 unique
cross-dimensional frame combinations. The most common single combination
(Threat: Agree, Utility: Agree, Transparency: Agree, Ownership: Artist
owns, Compensation: Would donate freely) accounts for just 18 artists
(7.1\%).

\subsection{Compression is invariant to artist AI
exposure}\label{compression-is-invariant-to-artist-ai-exposure}

As a robustness check on the structural compression claim made in
Results, we tested whether artists' self-reported AI exposure predicts
where their probes land in the 20-topic discourse space. Lovato et
al.~(2024) collected two exposure variables: (1) direct use of AI art
models (`Used\_AI\_art\_models') with values Yes / No / Don't Know /
None of the Above and (2) a categorical familiarity variable
(`AI\_models\_familiarity') with five levels None / AI art generally /
AI DeepLearning or GANS / text-to-image / LLMs.

We computed \(\chi^{2}\) contingency tests between each pair of exposure
subgroups across the 20 discourse topics. No comparison reached
statistical significance at \(\alpha\)= 0.05, and
Cramér\textquotesingle s V is uniformly small (all V \(\leq\) 0.081).

\subsection{Table S5: Topic-landing
distribution}\label{table-s5-topic-landing-distribution}

Topic-landing distribution does not differ between artist AI-exposure
subgroups (six pairwise \(\chi^{2}\) tests).

\begin{longtable}[]{@{}
  >{\raggedright\arraybackslash}p{(\linewidth - 10\tabcolsep) * \real{0.4306}}
  >{\raggedright\arraybackslash}p{(\linewidth - 10\tabcolsep) * \real{0.1181}}
  >{\raggedright\arraybackslash}p{(\linewidth - 10\tabcolsep) * \real{0.0972}}
  >{\raggedright\arraybackslash}p{(\linewidth - 10\tabcolsep) * \real{0.0625}}
  >{\raggedright\arraybackslash}p{(\linewidth - 10\tabcolsep) * \real{0.1389}}
  >{\raggedright\arraybackslash}p{(\linewidth - 10\tabcolsep) * \real{0.1389}}@{}}
\toprule\noalign{}
\begin{minipage}[b]{\linewidth}\raggedright
\textbf{Comparison}
\end{minipage} & \begin{minipage}[b]{\linewidth}\raggedright
\textbf{n (probes)}
\end{minipage} & \begin{minipage}[b]{\linewidth}\raggedright
\(\chi^{2}\)
\end{minipage} & \begin{minipage}[b]{\linewidth}\raggedright
\textbf{df}
\end{minipage} & \begin{minipage}[b]{\linewidth}\raggedright
\textbf{P (two-sided)}
\end{minipage} & \begin{minipage}[b]{\linewidth}\raggedright
\textbf{Cramér\textquotesingle s V}
\end{minipage} \\
\midrule\noalign{}
\endhead
\bottomrule\noalign{}
\endlastfoot
Used = Yes (148 resp) vs No (98 resp) & 1,229 & 4.90 & 4 & 0.30 &
0.063 \\
Familiarity: general (219) vs technical, GANs/T2I/LLMs (25) & 1,219 &
4.68 & 4 & 0.32 & 0.062 \\
Familiarity: none (8) vs any (244) & 1,259 & 4.02 & 4 & 0.40 & 0.056 \\
Combined: minimal/no use (105) vs technical+used (14) & 595 & 3.91 & 4 &
0.42 & 0.081 \\
Combined: minimal/no use (105) vs general use (133) & 1,189 & 4.70 & 4 &
0.32 & 0.063 \\
Combined: general use (133) vs technical+used (14) & 734 & 4.46 & 3 &
0.22 & 0.078 \\
\end{longtable}

All tests two-sided, \(\alpha\) = 0.05. n\_a\_respondents and
n\_b\_respondents counted at the respondent level; n (probes) is the
total contingency table sum. Underlying per-comparison data are in
figures/final\_pipeline/exposure\_subgroup\_analysis.csv.

Ownership and utility achieve total compression (100\% landing in T13
``AI Art Authenticity and Human Creativity'') at every exposure level,
and the four-topic top-k concentration exceeds 99.8\% in every subgroup
tested. We also note a small, non-significant directional trend, where
the T13 share rises with increasing exposure in transparency (16.2\%
\(\rightarrow\) 20.3\% \(\rightarrow\) 28.6\%) and threat (15.2\%
\(\rightarrow\) 26.5\% \(\rightarrow\) 28.6\%, across minimal / general
/ technical tiers respectively), while compensation rises only at the
technical tier (14.3\% \(\rightarrow\) 13.5\% \(\rightarrow\) 21.4\%).
The direction is consistent with the voice-collapse mechanism reported
in the main text. Technical AI familiarity may produce phrasing the
embedding model reads as more art-philosophical. It should be noted that
the small sample sizes in the technical-familiarity cells (n = 14
respondents in the technical+used tier; n = 3 in the LLMs-only
familiarity cell) preclude a confirmatory test. We flag the pattern for
follow-up with larger samples and note that the central claim of the
manuscript (structural compression of stakeholder concerns in public
AI-art discourse) does not depend on artist-level exposure.

\subsection{Table S6: Within-Theme Topic Distribution
Analysis}\label{table-s6-within-theme-topic-distribution-analysis}

To examine how positions within each concern dimension distribute across
discourse topics, we projected each artist probe into the consensus
space and recorded its nearest topic assignment. The results reveal
several distinct compression mechanisms, elaborated in the following
subsections.

\subsubsection{Five topics absorb all artist
voices}\label{five-topics-absorb-all-artist-voices}

Of the 20 public discourse topics, only 5 contain any artist probes:

\begin{longtable}[]{@{}
  >{\raggedright\arraybackslash}p{(\linewidth - 6\tabcolsep) * \real{0.1250}}
  >{\raggedright\arraybackslash}p{(\linewidth - 6\tabcolsep) * \real{0.4688}}
  >{\raggedright\arraybackslash}p{(\linewidth - 6\tabcolsep) * \real{0.2083}}
  >{\raggedright\arraybackslash}p{(\linewidth - 6\tabcolsep) * \real{0.1771}}@{}}
\toprule\noalign{}
\begin{minipage}[b]{\linewidth}\raggedright
\textbf{Topic}
\end{minipage} & \begin{minipage}[b]{\linewidth}\raggedright
\textbf{Label}
\end{minipage} & \begin{minipage}[b]{\linewidth}\raggedright
\textbf{Artist Probes}
\end{minipage} & \begin{minipage}[b]{\linewidth}\raggedright
\textbf{\% of Total}
\end{minipage} \\
\midrule\noalign{}
\endhead
\bottomrule\noalign{}
\endlastfoot
T13 & AI Art Authenticity and Human Creativity & 643 & 51.1\% \\
T1 & AI as Creative Collaborator & 551 & 43.8\% \\
T7 & Conversational Reflections on Art Practice & 56 & 4.4\% \\
T3 & Personal Reflections on AI and Art & 8 & 0.6\% \\
T19 & Generative Art History and Pioneers & 1 & 0.08\% \\
\end{longtable}

\subsubsection{The orphan rights topics (zero artist
probes)}\label{the-orphan-rights-topics-zero-artist-probes}

Three topics in the public discourse semantic map are interpreted by the
annotator panel as explicitly about artist rights, based on chunk
content (top keywords copyright, coders, software, law for T8;
Nightshade, Glaze, tools for T10; Midjourney, Allen, copyright for T18),
yet they contain zero artist probes:

\begin{longtable}[]{@{}
  >{\raggedright\arraybackslash}p{(\linewidth - 6\tabcolsep) * \real{0.1250}}
  >{\raggedright\arraybackslash}p{(\linewidth - 6\tabcolsep) * \real{0.4375}}
  >{\raggedright\arraybackslash}p{(\linewidth - 6\tabcolsep) * \real{0.2083}}
  >{\raggedright\arraybackslash}p{(\linewidth - 6\tabcolsep) * \real{0.2083}}@{}}
\toprule\noalign{}
\begin{minipage}[b]{\linewidth}\raggedright
\textbf{Topic}
\end{minipage} & \begin{minipage}[b]{\linewidth}\raggedright
\textbf{Label}
\end{minipage} & \begin{minipage}[b]{\linewidth}\raggedright
\textbf{Public Chunks}
\end{minipage} & \begin{minipage}[b]{\linewidth}\raggedright
\textbf{Artist Probes}
\end{minipage} \\
\midrule\noalign{}
\endhead
\bottomrule\noalign{}
\endlastfoot
T8 & AI Copyright and Legal Protection & 68 & 0 \\
T10 & Artist Defense Tools Against AI & 10 & 0 \\
T18 & AI Art Authorship and Copyright Debates & 53 & 0 \\
\end{longtable}

These three topics contain 131 public discourse chunks discussing
copyright disputes, legal protection, and defensive technologies. Three
further zero-artist topics are read by the annotator panel as
artist-themed (AI Authorship and Creative Agency, T11, 116 chunks;
Artist Reflections on Technology, T16, 165 chunks; Artist-Centered AI
Design and Ethics, T17, 120 chunks). Across all six topics, 532 chunks
of public discourse that trained readers (four LLMs and two human
coders, working from the chunk contents) interpret as being about
artists or artist rights, yet zero artist probes from any concern
dimension land in any of them. Public discourse has built substantial
conversational infrastructure for talking about artists and artist legal
protection without using language that maps to how artists themselves
articulate those concerns.

\subsubsection{Position-by-topic breakdown for each
theme}\label{position-by-topic-breakdown-for-each-theme}

\textbf{Threat (3 positions, 2 topics, binary split):}

\begin{longtable}[]{@{}
  >{\raggedright\arraybackslash}p{(\linewidth - 6\tabcolsep) * \real{0.2747}}
  >{\raggedright\arraybackslash}p{(\linewidth - 6\tabcolsep) * \real{0.0879}}
  >{\raggedright\arraybackslash}p{(\linewidth - 6\tabcolsep) * \real{0.0879}}
  >{\raggedright\arraybackslash}p{(\linewidth - 6\tabcolsep) * \real{0.5275}}@{}}
\toprule\noalign{}
\begin{minipage}[b]{\linewidth}\raggedright
\textbf{Position}
\end{minipage} & \begin{minipage}[b]{\linewidth}\raggedright
\textbf{n}
\end{minipage} & \begin{minipage}[b]{\linewidth}\raggedright
\textbf{\%}
\end{minipage} & \begin{minipage}[b]{\linewidth}\raggedright
\textbf{Destination topic}
\end{minipage} \\
\midrule\noalign{}
\endhead
\bottomrule\noalign{}
\endlastfoot
Agree (AI is a threat) & 152 & 60.6\% & T1: AI as Creative
Collaborator \\
Unsure & 44 & 17.5\% & T1: AI as Creative Collaborator \\
Disagree & 55 & 21.9\% & T13: AI Art Authenticity and Human
Creativity \\
\end{longtable}

The ``agree threat'' and ``unsure'' positions both land in a topic
labelled ``AI as Creative Collaborator,'' a positive framing of AI. The
``disagree threat'' position lands in a topic about aesthetic
authenticity. Neither destination preserves the labour argument the
threat statement makes. The discourse offers no semantic space for
substantive labour or economic claims about AI's role.

\textbf{Utility (3 positions, 1 topic, total compression):}

\begin{longtable}[]{@{}
  >{\raggedright\arraybackslash}p{(\linewidth - 6\tabcolsep) * \real{0.2111}}
  >{\raggedright\arraybackslash}p{(\linewidth - 6\tabcolsep) * \real{0.1000}}
  >{\raggedright\arraybackslash}p{(\linewidth - 6\tabcolsep) * \real{0.1333}}
  >{\raggedright\arraybackslash}p{(\linewidth - 6\tabcolsep) * \real{0.5333}}@{}}
\toprule\noalign{}
\begin{minipage}[b]{\linewidth}\raggedright
\textbf{Position}
\end{minipage} & \begin{minipage}[b]{\linewidth}\raggedright
\textbf{n}
\end{minipage} & \begin{minipage}[b]{\linewidth}\raggedright
\textbf{\%}
\end{minipage} & \begin{minipage}[b]{\linewidth}\raggedright
\textbf{Destination topic}
\end{minipage} \\
\midrule\noalign{}
\endhead
\bottomrule\noalign{}
\endlastfoot
Agree (positive) & 116 & 46.0\% & T13: AI Art Authenticity and Human
Creativity \\
Disagree & 69 & 27.4\% & T13: AI Art Authenticity and Human
Creativity \\
Unsure & 67 & 26.6\% & T13: AI Art Authenticity and Human Creativity \\
\end{longtable}

All three utility positions land in the same topic. T13 absorbs every
utility probe regardless of stance. Mutually opposite positions on
whether AI is a positive development for art group by register rather
than by substantive content. This is the cleanest single instance of
voice collapse in the study.

\textbf{Transparency (3 positions, 2 topics, near-total compression):}

\begin{longtable}[]{@{}
  >{\raggedright\arraybackslash}p{(\linewidth - 6\tabcolsep) * \real{0.2990}}
  >{\raggedright\arraybackslash}p{(\linewidth - 6\tabcolsep) * \real{0.0825}}
  >{\raggedright\arraybackslash}p{(\linewidth - 6\tabcolsep) * \real{0.1031}}
  >{\raggedright\arraybackslash}p{(\linewidth - 6\tabcolsep) * \real{0.4948}}@{}}
\toprule\noalign{}
\begin{minipage}[b]{\linewidth}\raggedright
\textbf{Position}
\end{minipage} & \begin{minipage}[b]{\linewidth}\raggedright
\textbf{n}
\end{minipage} & \begin{minipage}[b]{\linewidth}\raggedright
\textbf{\%}
\end{minipage} & \begin{minipage}[b]{\linewidth}\raggedright
\textbf{Destination topic}
\end{minipage} \\
\midrule\noalign{}
\endhead
\bottomrule\noalign{}
\endlastfoot
Agree (require disclosure) & 204 & 81.0\% & T1: AI as Creative
Collaborator \\
Unsure & 25 & 9.9\% & T13: AI Art Authenticity and Human Creativity \\
Disagree & 23 & 9.1\% & T13: AI Art Authenticity and Human Creativity \\
\end{longtable}

204 of 252 transparency probes (81\%) land in T1. The remaining 48 land
in T13. Neither topic's label captures the regulatory demand at the
heart of the transparency claim. Normalised entropy is 0.137.

\textbf{Ownership (24 positions, 1 topic, total compression):}

\begin{longtable}[]{@{}
  >{\raggedright\arraybackslash}p{(\linewidth - 4\tabcolsep) * \real{0.5106}}
  >{\raggedright\arraybackslash}p{(\linewidth - 4\tabcolsep) * \real{0.1915}}
  >{\raggedright\arraybackslash}p{(\linewidth - 4\tabcolsep) * \real{0.2766}}@{}}
\toprule\noalign{}
\begin{minipage}[b]{\linewidth}\raggedright
\textbf{Topic}
\end{minipage} & \begin{minipage}[b]{\linewidth}\raggedright
\textbf{n}
\end{minipage} & \begin{minipage}[b]{\linewidth}\raggedright
\textbf{\%}
\end{minipage} \\
\midrule\noalign{}
\endhead
\bottomrule\noalign{}
\endlastfoot
T13: AI Art Authenticity and Human Creativity & 252 & 100.0\% \\
\end{longtable}

All 24 distinct ownership opinion frames, spanning views from ``the
artist should own everything'' to ``nobody should own AI art'' to
position-dependent claims, collapse into a single discourse topic. The
topic's label points to a philosophical debate about authenticity rather
than property rights. This is the most extreme instance of frame
redirection in the study (FCR = 24).

\textbf{Compensation (37 positions, 5 topics):}

\begin{longtable}[]{@{}
  >{\raggedright\arraybackslash}p{(\linewidth - 4\tabcolsep) * \real{0.5976}}
  >{\raggedright\arraybackslash}p{(\linewidth - 4\tabcolsep) * \real{0.1829}}
  >{\raggedright\arraybackslash}p{(\linewidth - 4\tabcolsep) * \real{0.1951}}@{}}
\toprule\noalign{}
\begin{minipage}[b]{\linewidth}\raggedright
\textbf{Topic}
\end{minipage} & \begin{minipage}[b]{\linewidth}\raggedright
\textbf{n}
\end{minipage} & \begin{minipage}[b]{\linewidth}\raggedright
\textbf{\%}
\end{minipage} \\
\midrule\noalign{}
\endhead
\bottomrule\noalign{}
\endlastfoot
T1: AI as Creative Collaborator & 151 & 59.9\% \\
T7: Conversational Reflections on Art Practice & 56 & 22.2\% \\
T13: AI Art Authenticity and Human Creativity & 36 & 14.3\% \\
T3: Personal Reflections on AI and Art & 8 & 3.2\% \\
T19: Generative Art History and Pioneers & 1 & 0.4\% \\
\end{longtable}

Compensation is the only theme that spreads across more than two topics.
Even so, 96.4\% of probes concentrate in three topics (T1, T7, T13), and
none of these destination labels reference rights, property, or labour
demands.

\subsection{LLM theme validation}\label{llm-theme-validation}

To validate the theme assignments of the 750 public probes, we
classified each probe against the five artist-concern themes using two
large language models, GPT-5.4-mini (OpenAI) and Claude Opus 4.6
(Anthropic), under a multi-label classification prompt. The prompt
allowed each sentence to receive more than one theme assignment,
reflecting the observation that public-discourse sentences routinely
entangle several themes at once. Theme definitions were drawn verbatim
from the Lovato et al.~(2024) survey Likert template wording, and six
artist-probe exemplars per theme (30 exemplars total) were provided in
context as few-shot demonstrations.

The primary metric is \textbf{multi-label recall}: the fraction of
probes where the extraction's assigned theme appears in the LLM's
predicted theme set. Multi-label recall is appropriate here because the
aim is not to test whether an LLM picks the same single label as our
extraction, but whether the LLM confirms our label as one defensible
interpretation of the sentence.

\subsubsection{Per-theme multi-label recall (N=150 per theme, 95\%
Wilson
CIs)}\label{per-theme-multi-label-recall-n150-per-theme-95-wilson-cis}

\begin{longtable}[]{@{}
  >{\raggedright\arraybackslash}p{(\linewidth - 8\tabcolsep) * \real{0.2932}}
  >{\raggedright\arraybackslash}p{(\linewidth - 8\tabcolsep) * \real{0.0602}}
  >{\raggedright\arraybackslash}p{(\linewidth - 8\tabcolsep) * \real{0.2030}}
  >{\raggedright\arraybackslash}p{(\linewidth - 8\tabcolsep) * \real{0.2030}}
  >{\raggedright\arraybackslash}p{(\linewidth - 8\tabcolsep) * \real{0.2256}}@{}}
\toprule\noalign{}
\begin{minipage}[b]{\linewidth}\raggedright
\textbf{Theme}
\end{minipage} & \begin{minipage}[b]{\linewidth}\raggedright
\textbf{N}
\end{minipage} & \begin{minipage}[b]{\linewidth}\raggedright
\textbf{GPT-5.4-mini}
\end{minipage} & \begin{minipage}[b]{\linewidth}\raggedright
\textbf{Claude Opus 4.6}
\end{minipage} & \begin{minipage}[b]{\linewidth}\raggedright
\textbf{Inter-LLM \(\kappa\) (per-theme)}
\end{minipage} \\
\midrule\noalign{}
\endhead
\bottomrule\noalign{}
\endlastfoot
threat & 150 & 0.680 {[}0.602, 0.749{]} & 0.713 {[}0.636, 0.780{]} &
0.756 \\
utility & 150 & 0.827 {[}0.758, 0.879{]} & 0.747 {[}0.672, 0.810{]} &
0.768 \\
ownership & 150 & 0.793 {[}0.722, 0.850{]} & 0.780 {[}0.707, 0.839{]} &
0.836 \\
transparency & 150 & 0.667 {[}0.588, 0.737{]} & 0.693 {[}0.615, 0.762{]}
& 0.831 \\
Four clean themes (mean) & 600 & 0.742 {[}0.705, 0.775{]} & 0.733
{[}0.697, 0.767{]} & 0.80 avg \\
\emph{compensation (analysed separately)} & \emph{150} & \emph{0.193
{[}0.138, 0.264{]}} & \emph{0.280 {[}0.214, 0.357{]}} & \emph{0.636} \\
\end{longtable}

\subsubsection{Inter-LLM agreement}\label{inter-llm-agreement}

Across all 3,750 probe-theme decisions, inter-LLM reliability was
Cohen's \textbf{\(\kappa\) = 0.795} (substantial agreement on the McHugh
2012 benchmarks: \(\kappa\) \textgreater{} 0.60 moderate, \(\kappa\)
\textgreater{} 0.80 strong). The two models produced \textbf{identical
multi-label sets for 70.4\% of probes} and shared at least one label for
94.3\%. Mean Jaccard similarity between the two models' predicted label
sets was 0.823 per probe. Inter-LLM agreement is therefore substantially
higher than either LLM's marginal agreement with the extraction itself
(\textasciitilde0.74), a pattern consistent with the two models
converging on coherent multi-label interpretations of the corpus.

\subsubsection{Compensation as an edge
case}\label{compensation-as-an-edge-case}

Compensation is the one theme where extraction and LLM labelling
systematically diverge. Of the 150 compensation-extracted probes, only
\textbf{27 (18.0\%) were confirmed by both LLMs} as compensation; 106
(70.7\%) were rejected by both. For the 106 rejected probes, the two
LLMs converged on identical alternative themes at a high rate:

\begin{longtable}[]{@{}
  >{\raggedright\arraybackslash}p{(\linewidth - 4\tabcolsep) * \real{0.4940}}
  >{\raggedright\arraybackslash}p{(\linewidth - 4\tabcolsep) * \real{0.2410}}
  >{\raggedright\arraybackslash}p{(\linewidth - 4\tabcolsep) * \real{0.2410}}@{}}
\toprule\noalign{}
\begin{minipage}[b]{\linewidth}\raggedright
\textbf{Alternative theme selected by LLMs}
\end{minipage} & \begin{minipage}[b]{\linewidth}\raggedright
\textbf{Opus 4.6}
\end{minipage} & \begin{minipage}[b]{\linewidth}\raggedright
\textbf{GPT-5.4-mini}
\end{minipage} \\
\midrule\noalign{}
\endhead
\bottomrule\noalign{}
\endlastfoot
ownership (copyright, rights, IP) & 46.2\% & 43.4\% \\
utility (creative economy, benefit) & 39.6\% & 50.0\% \\
threat (income loss, undercutting) & 29.2\% & 18.9\% \\
transparency (disclosure, consent) & 13.2\% & 15.1\% \\
\end{longtable}

This pattern is not a validation failure. It is a direct measurement of
the frame-redirection mechanism documented in the main text. Public
discourse routinely expresses compensation concerns through copyright
language (``protecting artists' rights over their training data''),
creative-economy language (``undermining artists' livelihoods''), and
threat language (``undercutting creative work''), rather than through
direct payment language. The two LLMs, independently evaluating each
probe, identify these as ownership, utility, and threat claims, exactly
as the semantic geometry of public discourse routes them in our
clustering analysis. Put differently: the same compression phenomenon
the manuscript documents in the clustering step also shows up in the LLM
validation step, at the sentence level.

Full per-probe classifications (both models, multi-label outputs),
per-theme confidence intervals, and the computed kappa values are
available in the project repository under
figures/final\_pipeline/approach\_f\_per\_probe\_full.csv and
figures/final\_pipeline/approach\_f\_recall\_with\_cis.csv.

\subsection{Consensus UMAP implementation
details}\label{consensus-umap-implementation-details}

\includegraphics[width=6.5in,height=3.54167in]{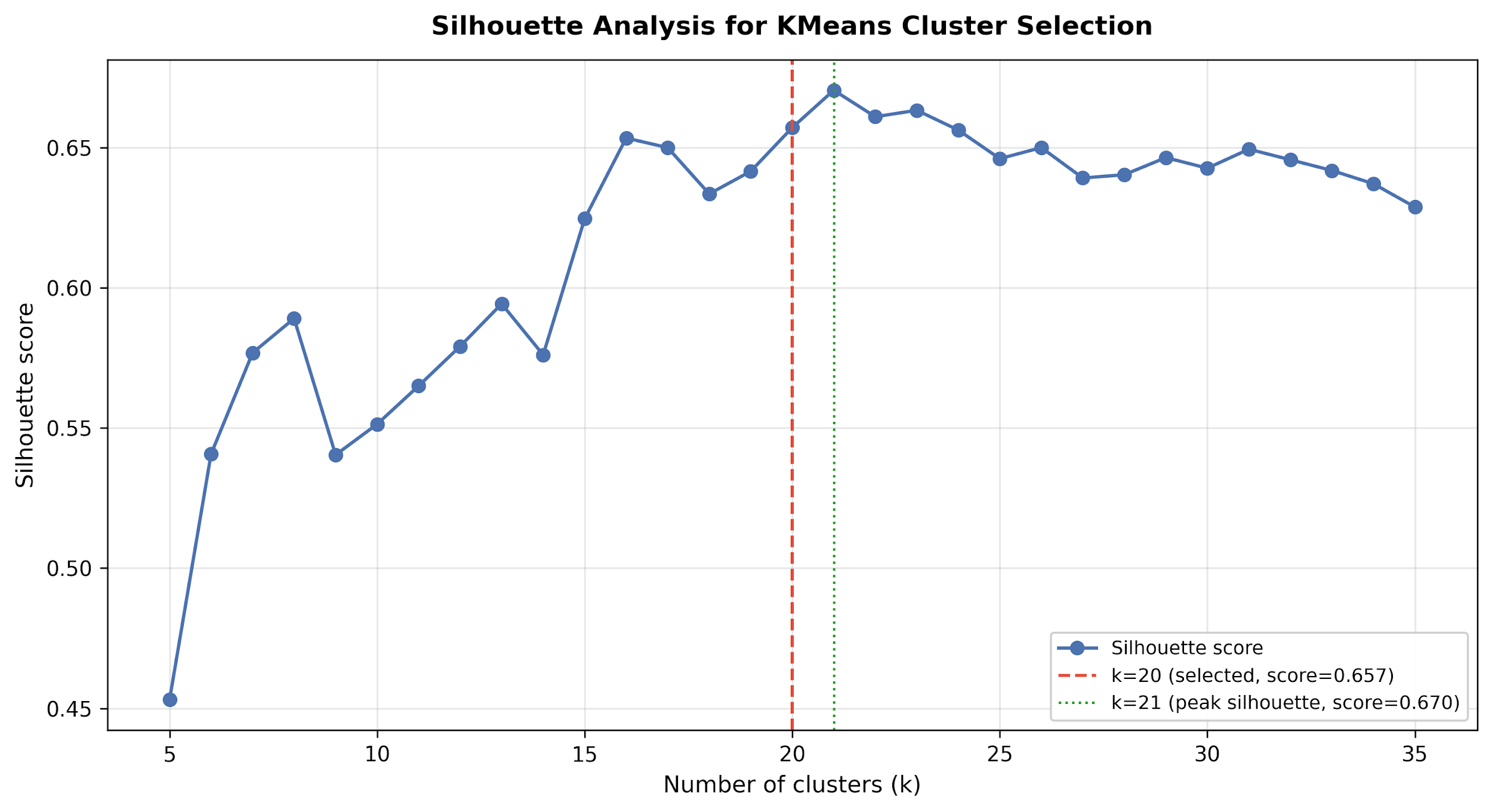}

\emph{\textbf{Figure S3.} Silhouette analysis for KMeans topic
selection. Scores computed via sklearn.metrics.silhouette\_score
(Euclidean) across k=5 to k=35 on the 5-dimensional consensus UMAP
coordinates. The silhouette curve peaks at k=21 (0.670), with k=20
scoring 0.657. k=20 was selected via the combined four-stage criteria
(consensus silhouette, ARI between per-seed and consensus partitions,
valid-topic rate, and single-article topic count) rather than silhouette
alone, because at k=21 the marginal silhouette gain (\(\Delta\) =
+0.013) does not offset the loss in valid-topic rate. Sensitivity
analysis confirms that the top-4 concentration of artist probes exceeds
91\% at all tested k values (k=10 to 30).}

\subsubsection{Parameters and
configuration}\label{parameters-and-configuration}

\begin{longtable}[]{@{}
  >{\raggedright\arraybackslash}p{(\linewidth - 2\tabcolsep) * \real{0.0600}}
  >{\raggedright\arraybackslash}p{(\linewidth - 2\tabcolsep) * \real{0.9333}}@{}}
\toprule\noalign{}
\begin{minipage}[b]{\linewidth}\raggedright
\textbf{Parameter}
\end{minipage} & \begin{minipage}[b]{\linewidth}\raggedright
\textbf{Value}
\end{minipage} \\
\midrule\noalign{}
\endhead
\bottomrule\noalign{}
\endlastfoot
Embedding model & e5-large-v2 with ``query:'' prefix \\
Corpus & 1,736 chunks from 125 articles \\
n\_components & 5 \\
n\_neighbors & 53 \\
min\_dist & 0.01 \\
metric & cosine \\
Number of seeds & 30 \\
Clustering & KMeans, k=20 \\
Projection head & scikit-learn MLPRegressor; hidden layers (1024, 512,
256, 128), ReLU, L2 \(\alpha\)=0.0001, Adam with initial learning rate
0.002, max 1000 iters, early stopping on 10\% inner validation split,
StandardScaler on X and Y, 85/15 train/test split, random\_state=42,
R\textsuperscript{2}=0.904 on held-out test \\
\end{longtable}

Seeds: {[}137, 85, 127, 59, 195, 243, 170, 77, 186, 79, 69, 42, 240,
105, 199, 91, 151, 82, 177, 234, 46, 101, 34, 175, 108, 81, 176, 241,
20, 53{]}.

Parameters selected via 4-stage grid search (497 configurations). Config
A (nn=53, md=0.01, nc=5) chosen for highest consensus silhouette
(0.657), fewest single-article topics (3 of 20), and highest valid topic
rate (85\%). At k=20 the consensus silhouette is 0.657 (computed via
sklearn.metrics.silhouette\_score on the 5-dimensional consensus UMAP
coordinates with Euclidean distance); the silhouette curve across k =
5--35 (Figure S3) peaks at k=21 (0.670). k=20 was preferred because the
marginal silhouette gain at k=21 (\(\Delta\) = +0.013) does not offset
the loss in valid-topic rate. We report the consensus silhouette here,
which is distinct from the per-seed silhouettes (silhouette of each
seed\textquotesingle s clustering in its own UMAP embedding) and from
the seed\(\leftrightarrow\)consensus ARI (label-partition agreement, not
a silhouette).

\subsubsection{Distance-matrix consensus
method}\label{distance-matrix-consensus-method}

For each seeded UMAP projection, the pairwise Euclidean distance matrix
in 5D space was computed. These 30 matrices were element-wise averaged
to produce a consensus distance matrix. A final UMAP was fit using this
consensus distance structure as the precomputed distance input.

\begin{longtable}[]{@{}
  >{\raggedright\arraybackslash}p{(\linewidth - 2\tabcolsep) * \real{0.5250}}
  >{\raggedright\arraybackslash}p{(\linewidth - 2\tabcolsep) * \real{0.4500}}@{}}
\toprule\noalign{}
\begin{minipage}[b]{\linewidth}\raggedright
\textbf{Method}
\end{minipage} & \begin{minipage}[b]{\linewidth}\raggedright
\textbf{Mean ARI (seed vs.~consensus)}
\end{minipage} \\
\midrule\noalign{}
\endhead
\bottomrule\noalign{}
\endlastfoot
Naive coordinate averaging (Procrustes) & 0.56 \\
Distance-matrix consensus & 0.71 \\
\end{longtable}

\subsubsection{Style-control comparison
(H2)}\label{style-control-comparison-h2}

Quantitative summary of the style-control test reported in §H2 of
Results. Raw public discourse refers to the 1,736 chunked media corpus;
style-matched public probes refers to the 750 retrieved short
declarative sentences (extracted as described below); artist refers to
the 1,259 templated artist probes from the Lovato et al.~survey.

\begin{longtable}[]{@{}
  >{\raggedright\arraybackslash}p{(\linewidth - 6\tabcolsep) * \real{0.3750}}
  >{\raggedright\arraybackslash}p{(\linewidth - 6\tabcolsep) * \real{0.1875}}
  >{\raggedright\arraybackslash}p{(\linewidth - 6\tabcolsep) * \real{0.1607}}
  >{\raggedright\arraybackslash}p{(\linewidth - 6\tabcolsep) * \real{0.2589}}@{}}
\toprule\noalign{}
\begin{minipage}[b]{\linewidth}\raggedright
\textbf{Comparison}
\end{minipage} & \begin{minipage}[b]{\linewidth}\raggedright
\textbf{Jensen-Shannon}
\end{minipage} & \begin{minipage}[b]{\linewidth}\raggedright
\textbf{Cramér\textquotesingle s V}
\end{minipage} & \begin{minipage}[b]{\linewidth}\raggedright
\textbf{Centroid Distance (5D)}
\end{minipage} \\
\midrule\noalign{}
\endhead
\bottomrule\noalign{}
\endlastfoot
Public discourse vs Artist (raw) & 0.363 & 0.740 & 2.003 \\
Public probes vs Artist (style-matched) & 0.308 & 0.734 & 3.754 \\
\end{longtable}

Source: figures/final\_pipeline/all\_metrics.csv (h2\_jsd\_raw,
h2\_jsd\_style, h2\_cramers\_v\_raw, h2\_cramers\_v\_style,
h2\_centroid\_raw, h2\_centroid\_style). Distances are Euclidean on the
5-dimensional consensus UMAP coordinates. Style controls reduce JSD by
15.4\% (0.364 \(\rightarrow\) 0.308) but leave Cramér\textquotesingle s
V essentially unchanged (1\% reduction) and increase the centroid
distance from 2.003 to 3.754. Approximately 84.6\% of the raw divergence
persists after style control and is therefore semantic rather than
stylistic.

\subsubsection{Public probe extraction}\label{public-probe-extraction}

750 style-matched public probes were extracted using embedding-based
retrieval. 250 synthetic Likert anchors (5 themes \(\times\) 5 agreement
levels \(\times\) 10 discourse styles) were embedded and used as queries
to retrieve nearest-neighbour sentences from the 1,736-chunk corpus by
cosine similarity. Candidates were then filtered to retain only on-theme
sentences and deduplicated per theme, yielding the final 750-probe set
used for all style-matched comparisons in the main analysis.

\subsubsection{Free-text validation}\label{free-text-validation}

38 organic free-text responses from the Lovato et al.~survey were
embedded and projected into the consensus space. 84\% of these responses
land in the same topic region as the corresponding templated
compensation probes.

\subsubsection{2D visualisation}\label{d-visualisation}

For 2D visualisation, PCA is applied to the 5-dimensional consensus
coordinates. PCA captures 80.8\% of the variance in the first two
components (PC1: 66.6\%, PC2: 14.2\%).

\includegraphics[width=6.5in,height=2.70833in]{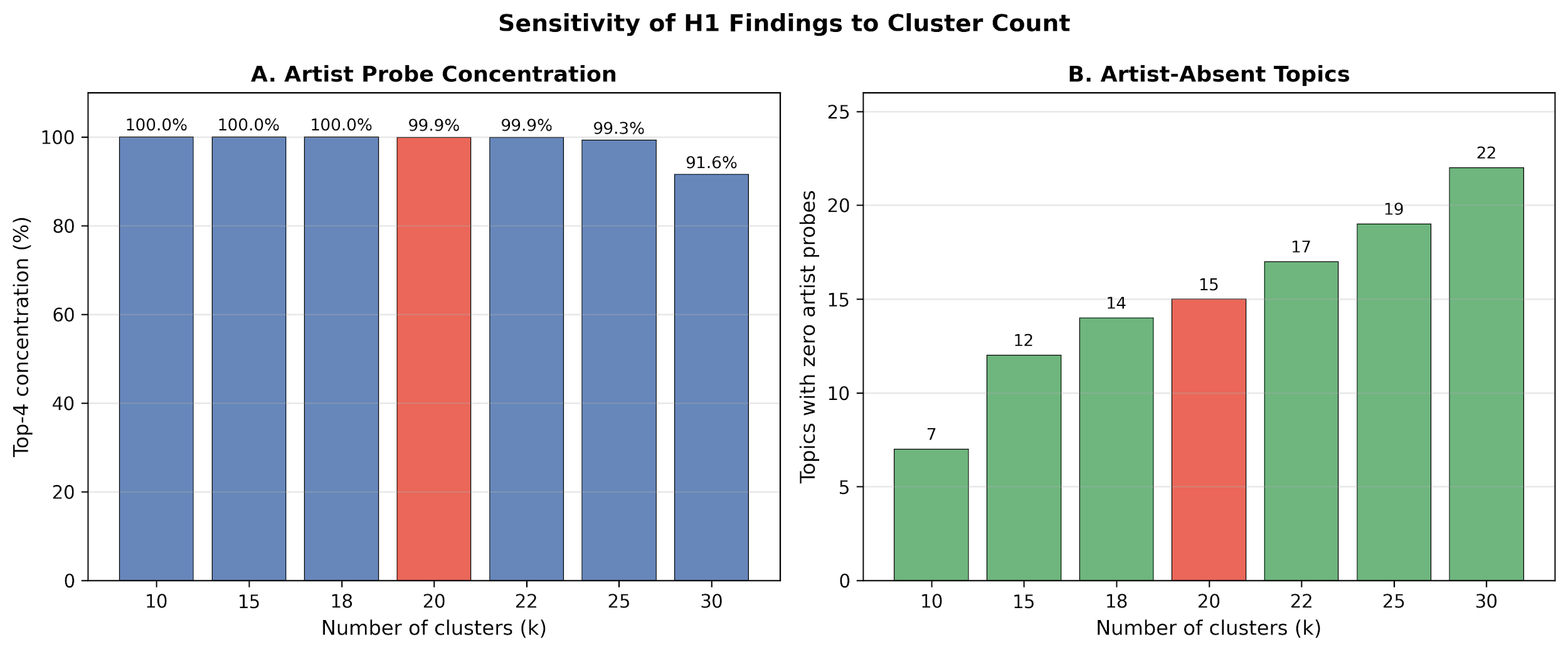}

\emph{\textbf{Figure S4.} Sensitivity of the artist-concentration
finding to topic count. (A) Top-4 topic concentration of artist probes
remains above 91\% across all k values from 10 to 30. (B) The number of
topics with zero artist probes increases with k, as expected, but the
pattern of extreme concentration is consistent. The robustness of both
patterns across cluster counts indicates that the topical-exclusion
finding is structural rather than an artifact of the k=20 choice.}

\end{document}